\pdfoutput=1
\documentclass[11pt]{article}

\usepackage[preprint]{acl}
\usepackage[most]{tcolorbox} % For creating stylish boxes
\usepackage{times}
\usepackage{latexsym}

\usepackage[T1]{fontenc}
\usepackage[utf8]{inputenc}

\usepackage{microtype}

\usepackage{inconsolata}

\usepackage{graphicx}

\usepackage{multirow}
\usepackage{adjustbox}
\usepackage{array}

\usepackage{booktabs}  
\usepackage{multirow}
\usepackage{graphicx}
\usepackage{subcaption}
\usepackage{url}
\usepackage{enumitem}
\usepackage{wrapfig}
\usepackage[utf8]{inputenc} % allow utf-8 input
\usepackage[T1]{fontenc}    % use 8-bit T1 fonts
\usepackage{hyperref}       % hyperlinks
\usepackage{url}            % simple URL typesetting
\usepackage{amsfonts}       % blackboard math symbols
\usepackage{nicefrac}       % compact symbols for 1/2, etc.
\usepackage{microtype}      % microtypography
\usepackage{xcolor}         % colors
\usepackage{natbib}
\usepackage{caption}
\usepackage{multirow}
\usepackage{graphicx}
\usepackage{physics}
\usepackage{xr-hyper}
\usepackage{url}            % simple URL typesetting
\usepackage{amsfonts}       % blackboard math symbols
\usepackage{nicefrac} 
\usepackage{microtype}      % microtypography
\usepackage{graphicx}
\usepackage{booktabs} % for professional tables
\usepackage{amsfonts}
\usepackage{amssymb,bm}
\usepackage{amstext, enumitem}
\usepackage{amsmath}
\usepackage{xspace}
\usepackage{amsthm}
\usepackage{algorithm}
\usepackage{algpseudocode}
\usepackage{graphicx,color}
\usepackage{url}
\usepackage{graphics}
\usepackage{colordvi, wrapfig}
\usepackage{colordvi}
\usepackage{makecell}
\usepackage{todonotes}
\usepackage{amsthm}
\newtheorem{definition}{Definition}[section]
\newtheorem{assumption}{Assumption}[section]
\newtheorem{theorem}{Theorem}[section]
\newtheorem{lemma}{Lemma}[section]
\newtheorem{remark}{Remark}

\title{From Parameters to Prompts: Understanding and Mitigating the Factuality Gap between Fine-Tuned LLMs}

\author{
  Xuan Gong$^{1}$ \quad Hanbo Huang$^{2}$ \quad Shiyu Liang$^{2}$\thanks{Corresponding author} \\
  $^{1}$Tongji University \quad $^{2}$Shanghai Jiao Tong University \\
  \texttt{2152095@tongji.edu.cn} \quad 
  \texttt{\{hhuang417, lsy18602808513\}@sjtu.edu.cn}
}

\begin{document}
\maketitle
\begin{abstract}
Factual knowledge extraction aims to explicitly extract knowledge parameterized in pre-trained language models for application in downstream tasks. While prior work has been investigating the impact of supervised fine-tuning data on the factuality of large language models (LLMs), its mechanism remains poorly understood. We revisit this impact through systematic experiments, with a particular focus on the factuality gap that arises when fine-tuning on known versus unknown knowledge. Our findings show that this gap can be mitigated at the inference stage, either under out-of-distribution (OOD) settings or by using appropriate in-context learning (ICL) prompts (i.e., few-shot learning and Chain of Thought (CoT)). We prove this phenomenon theoretically from the perspective of knowledge graphs, showing that the test-time prompt may diminish or even overshadow the impact of fine-tuning data and play a dominant role in knowledge extraction. Ultimately, our results shed light on the interaction between finetuning data and test-time prompt, demonstrating that ICL can effectively compensate for shortcomings in fine-tuning data, and highlighting the need to reconsider the use of ICL prompting as a means to evaluate the effectiveness of fine-tuning data selection methods.  

\end{abstract}

\section{Introduction}

Pre-trained large language models (LLMs) store extensive parameterized knowledge~\citep{meng2022locating,petroni-etal-2019-language,allenzhu2024physicslanguagemodels31}, which can be extracted and applied to various downstream tasks through different prompt designs~\citep{chen2024unleashingpotentialpromptengineering,wang2024retaskrevisitingllmtasks}. However, querying LLMs with naturally phrased questions may increase the likelihood of generating incorrect answers, leading to model hallucinations~\citep{zhang-etal-2024-clamber,hallucination_survey}. Previous research has shown that fine-tuning LLMs can enhance their factuality~\citep{51119}, yet the impact varies significantly depending on the dataset. For instance, \citet{gekhman-etal-2024-fine} and \citet{ghosal2024understanding} indicate that fine-tuning on well-established or popular knowledge improves model performance, while fine-tuning on unknown or unpopular data can have the opposite effect.

%overview picture
\begin{figure}[t]
    \centering
    \includegraphics[width=\linewidth]{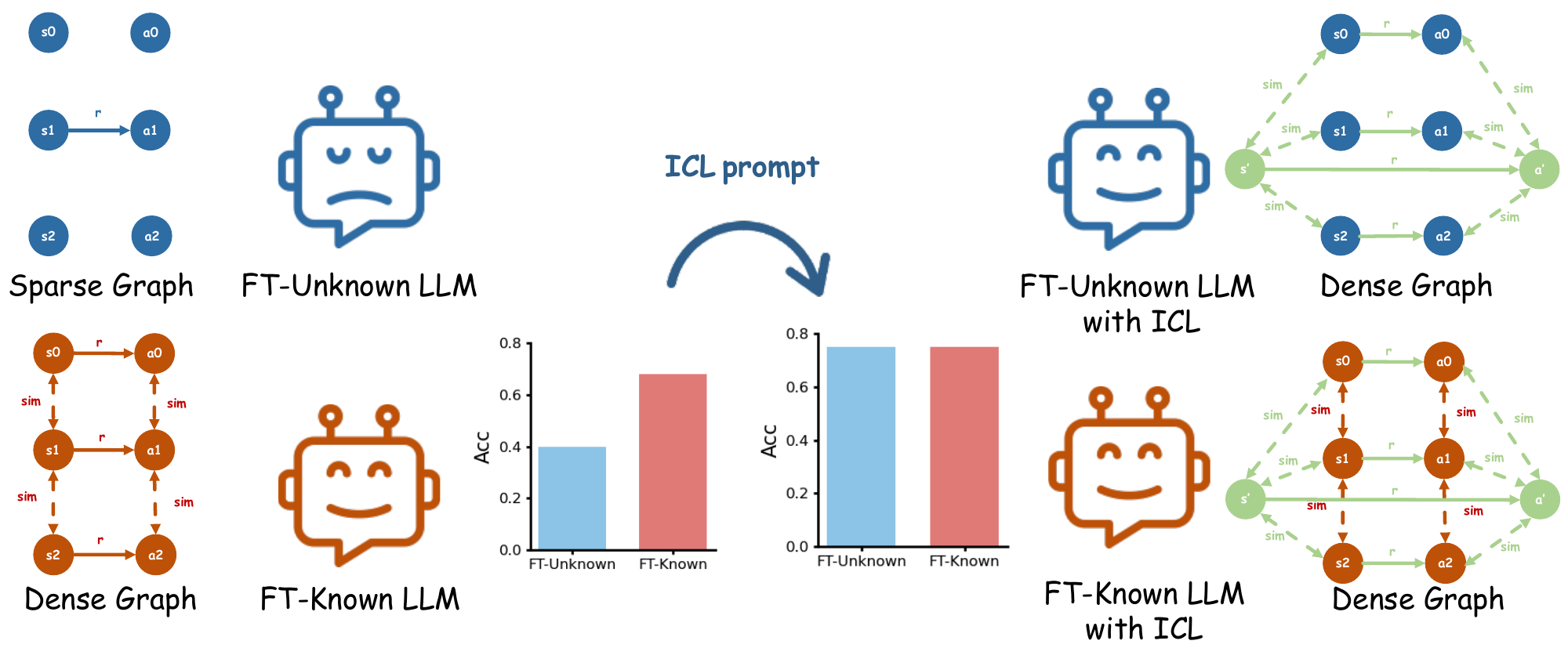}
    \caption{Overview: In-context learning (ICL) prompts can help reduce the factuality gap, as they enhance the connectivity of the graph of the FT-Unknown LLM by incorporating demonstrations like $(s', a')$, thereby narrowing the factuality gap. FT-Unknown LLM and FT-Known LLM refer to LLM fine-tuned on unknown and known knowledge, respectively.}
    \label{fig:overview}
    \vspace{-0.3cm}
\end{figure}

Previous research has extensively explored how different fine-tuning datasets impact the factuality of LLMs\citep{gekhman-etal-2024-fine,kazemi2023understandingfinetuningfactualknowledge,joshi-etal-2024-personas,ghosal2024understanding}. In this work, however,  we find that this factuality gap caused by finetuning data is highly fragile. Modifying the test-time prompt, such as through few-shot examples \citep{brown2020languagemodelsfewshotlearners} or chain-of-thought (CoT) \cite{NEURIPS2022_cot}, can significantly reduce or even reverse the gap. Our work suggests that the factuality gap caused by fine-tuning data can be understood from a novel perspective of knowledge graph modeling.

To gain deeper insight into the nature of this factuality gap, we pose the following three intriguing research questions: 
\textbf{RQ1:} \textit{How to understand the factuality gap caused by finetuning data?}
\textbf{RQ2:} \textit{Can the factuality gap be easily mitigated?}
\textbf{RQ3:} \textit{What can we do to utilize this finding in knowledge extraction?}
We select two types of models, the Llama-3.1-8B \citep{dubey2024llama} and Mistral-7B-v0.3 \citep{jiang2023mistral7b}, in both their \textit{Base} and \textit{Instruct} versions, and conduct experiments on two task categories: question answering (QA) and open-ended generation. These experiments allow us to answer the above questions. In this paper, our main contributions can be summarized as follows:
%[leftmargin=*, itemsep=0pt, parsep=0pt, topsep=0pt]
\begin{itemize}[leftmargin=*, topsep=4pt]
    \item Through extensive experiments, we validate the existence of a factuality gap introduced by fine-tuning data and demonstrate that this gap diminishes as the distributional distance of the test set increases. Furthermore, we identify in-context learning at inference time as an effective approach to mitigate this gap. 

    \item We conduct an in-depth analysis of the factuality gap and offer a deeper understanding from the perspective of knowledge graphs. To the best of our knowledge, we are the first to prove this phenomenon theoretically through the lens of graph modeling.
    
    \item Building on our empirical and theoretical work, we leverage this finding to explore its potential applications, especially introduce novel insights into the evaluation of data selection algorithms. 
\end{itemize}

\section{Related Works}
\subsection{Factual Knowledge Extraction in LLM}
LLMs store extensive world knowledge within their parameters, and ineffective extraction is a major cause of model hallucinations~\citep{pmlr-v202-kandpal23a,mallen-etal-2023-trust}. Therefore, understanding knowledge extraction is crucial for improving LLM efficiency and performance. \citet{allenzhu2024physicslanguagemodels31} integrates pretraining and fine-tuning to highlight the importance of data augmentation for extractable knowledge. \citet{yin-etal-2024-benchmarking} introduces the concept of a \textit{knowledge boundary}, where knowledge that cannot be correctly accessed under any expression is considered outside the model’s boundary. While prior work focuses on either pretraining and fine-tuning phases or extraction during inference, we study the interaction between model fine-tuning and inference to offer a more comprehensive analysis of factual knowledge extraction.

\subsection{Finetuning Data and Model Factuality}
Recent studies have explored the impact of fine-tuning data on model factuality. \citet{kang2024unfamiliar} suggests that unfamiliar examples in the fine-tuning dataset affect how the model handles unfamiliar test instances, but they do not address how these examples influence the overall factuality of the model. \citet{gekhman-etal-2024-fine} empirically demonstrate that fine-tuning on unknown knowledge negatively impacts factuality, attributing this to overfitting on such data during training. \citet{ghosal2024understanding} shows that finetuning on lesser-known facts leads to worse factuality because of less attention on the entity tokens during training. \citet{lin2024flame, liu2024tsds} attempt to improve the factuality of the model by refining the data used for fine-tuning. Extending prior work, we examine the impact of fine-tuning data on model factuality from the graph modeling angle, and propose a method to reduce its adverse effects.

\subsection{In-context Learning and Model Factuality}
As a test-time method, ICL plays an important role in LLM knowledge extraction capabilities, and many studies have explored how ICL affects model factuality. Some works focus on using ICL for knowledge editing \citep{zheng-etal-2023-edit}, while others investigate how the construction of ICL examples influences knowledge extraction \citep{wang-etal-2024-knowledgeable, Wu_2025, yang2024supervised, lin2024the}. In contrast, our work steps further to study the impact of ICL prompts on the factuality of fine-tuned models, rather than improving the ICL construction itself.

\section{Preliminaries and Setup} \label{sec:pre}
\subsection{Factual Knowledge}
Following prior work on factual knowledge in LLMs~\cite{ghosal2024understanding, petroni2019languagemodelsknowledgebases}, we represent each factual statement as a triplet \( (s, r, a) \), where \( s \) is the subject entity, \( r \) is the relation type, and \( a \) is the answer. This triplet structure is widely used in benchmarks such as LAMA~\cite{petroni2019languagemodelsknowledgebases}, KILT~\cite{petroni-etal-2021-kilt}, and TruthfulQA~\cite{lin-etal-2022-truthfulqa}.
Formally, we denote a piece of knowledge as \( k = (s, r, a) \in \mathcal{S} \times \mathcal{R} \times \mathcal{A} \), where \( \mathcal{S} \), \( \mathcal{R} \), and \( \mathcal{A} \) represent the sets of all subject entities, relation types, and answers, respectively. This abstraction provides a unified format for evaluating whether a language model contains and can retrieve specific facts.

\subsection{Knowledge Extraction in LLMs}
To analyze the mechanism of knowledge extraction, we consider a simplified one-layer transformer architecture, with fixed non-orthogonal embeddings \(E\in \mathbb{R}^{|\mathcal{T}|\times d}\) and the vocabulary \( \mathcal{T}\). An input sequence  of \( n \) tokens is written as \( X = (x_1, \ldots, x_n) \in \mathcal{T} \). The model computes its outputs as
\begin{align*}
f(X; W^{KQ}, W^{V}) = \sigma(\text{Att}(E(X);W^{KQ},W^V) ) ,
\end{align*}
where \( W^{KQ}, W^V \in \mathbb{R}^{d \times d} \) are learnable parameters, \(\text{Att}()\) is the self-attention function and \(\sigma()\) is the function that predict next token from the probability distribution. In this paper, we focus on the prediction for the next token, given by the final output vector \( f_{[:, -1]}(X; W, V) \).  For more detailed model settings, please refer to Appendix~\ref{app:notation}. 

Given a factual triplet \( (s, r, a) \in \mathcal{T}^3 \), we ask whether the model can retrieve the answer \( a \) when provided with an appropriate context. Specifically, we allow any context sequence \( \{x_1, \ldots, x_{n-1}\} \in \mathcal{T} \setminus \{s, r, a\} \), with \( s \) and \( r \). If the model predicts \( a \) as the next token, we say that the knowledge has been successfully extracted.

\begin{definition}[Unknown Knowledge]
A token triple $(s,r,a) \in \mathcal{T}^3$ is said to be an  \textbf{unknown knowledge} if, for all contexts $\{x_1, \ldots, x_{n-1}\} \subset \mathcal{T} \setminus \{s,r,a\}$, $f_{[:, -1]}(x_1, \ldots, x_{n-1}, s,r) \neq a.$

\end{definition}

\begin{definition}[Known Knowledge]
\label{def:known}
A token triple $(s, r, a) \in \mathcal{T}^3$ is said to be a \textbf{known knowledge} if there exists a context $\{x_1, \ldots, x_{n-1}\} \subset \mathcal{T} \setminus \{s, r,a\}$ such that
$f_{[:, -1]}(x_1, \ldots, x_{n-1}, s,r) = a.$

\end{definition}
In practice, we approximate this distinction using few-shot prompting. A triplet is considered known if the model produces the correct answer in at least one prompt. Otherwise, it is treated as unknown. This empirical definition enables generalization across prompt templates while preserving alignment with the formal setting above.

\begin{table*}[t]
\centering
\begin{tabular}{cccccccccc}
\hline
\multicolumn{1}{c}{\multirow{2}{*}{Dataset}} & \multirow{2}{*}{Split} 
& \multicolumn{2}{c}{Llama} 
& \multicolumn{2}{c}{Llama-Instruct} 
& \multicolumn{2}{c}{Mistral} 
& \multicolumn{2}{c}{Mistral-Instruct} \\ \cline{3-10}
 & & ES & Con. & ES & Con. & ES & Con. & ES & Con. \\ \hline

\multirow{2}{*}{EQ} 
 & Unknown & 28.25 & 24.80 & 28.75 & 25.00 & 21.15 & 18.00 & 26.00 & 20.90 \\
 & Known   & 40.30 & 38.50 & 39.20 & 37.70 & 36.05 & 34.45 & 35.40 & 34.50 \\ \hline

\multirow{2}{*}{PopQA} 
 & Unknown & 31.28 & 26.98 & 30.09 & 27.33 & 26.94 & 20.54 & 25.26 & 19.59 \\
 & Known   & 36.81 & 35.55 & 35.86 & 35.09 & 33.00 & 31.67 & 32.26 & 31.63 \\ \hline

\multirow{2}{*}{MMLU} 
 & Unknown & 34.94 & 33.90 & 33.64 & 33.51 & 28.09 & 26.52 & 31.61 & 25.87 \\
 & Known   & 37.49 & 37.10 & 35.92 & 34.88 & 35.60 & 34.81 & 33.44 & 32.14 \\ \hline

\multirow{2}{*}{WikiBios} 
 & Unknown & 55.50 & 46.90 &   &   & 47.30 & 36.67 &   &  \\
 & Known   & 58.25 & 49.69 &  &  & 49.16 & 39.58 &  &  \\ \hline
 
\end{tabular}
\vspace{-0.2cm}
\caption{QA tasks exact match accuracy and WikiBios FActScore evaluation. ES: Early Stop, Con.: Convergence. Llama: Llama-3.1-8B, Llama-Instrcut: Llama-3.1-8B-Instruct, Mistral: Mistral-7B-v0.3, Mistral-Instruct: Mistral-7B-Instruct-v0.3}
\label{tab:qa}
\vspace{-0.3cm}
\end{table*}
\section{Understanding the Factuality Gap from Finetuning on Known vs Unknown Knowledge (RQ1)}\label{sec:racap}
In this section, we examine the factuality gap in models fine-tuned on known versus unknown knowledge, under both in-distribution and out-of-distribution scenarios. We present comprehensive experimental observations and support them with corresponding theoretical analysis.

\subsection{Factuality Gap under In-distribution Generalization}
\label{sec:racap1}
\textbf{Settings.}
We evaluate the impact of fine-tuning on known versus unknown factual knowledge across two task settings: QA and open-ended generation. For QA task, we follow the experimental protocol of~\citet{gekhman-etal-2024-fine}, fine-tuning both base and instruction-tuned variants of LLaMA3.1-8B~\footnote{https://huggingface.co/meta-llama/\{Llama-3.1-8B, Llama-3.1-8B-Instruct\}} and Mistral-7B-v0.3~\footnote{https://huggingface.co/mistralai/\{Mistral-7B-v0.3, Mistral-7B-Instruct-v0.3\}} on known and unknown subsets derived from EntityQuestions~\citep{sciavolino2021simple}, PopQA~\citep{mallen-etal-2023-trust}, and MMLU~\citep{hendrycks2020mmlu}. Exact match accuracy is used as the evaluation metric. For open-ended generation task, we follow~\citet{kang2024unfamiliar} using the WikiBios dataset~\citep{stranisci-etal-2023-wikibio}. The dataset is split analogously into known and unknown subsets, and performance is measured using the FActScore metric~\citep{min-etal-2023-factscore}. All models are evaluated under both early stopping and full convergence conditions. Implementation details  are provided in Appendix~\ref{app:exp}.

%TODO: 更新下表中POpQA的部分

% \textbf{Context: Instruction-tuned models exhibit a reduced factuality gap compared to base models, with the gap widening as training progresses from early stopping to convergence.}
% Table~\ref{tab:qa} presents evaluation results for both early stopping and full convergence checkpoints. As training progresses, we observe that the factuality gap between models fine-tuned on known versus unknown knowledge consistently widens. This trend reflects that longer optimization tends to reinforce memorization of accessible (known) knowledge, while models trained on unknown data struggle to generalize~\cite{???}. Notably, instruction-tuned models show a consistently smaller factuality gap compared to their base counterparts across both LLaMA and Mistral architectures. This observation is consistent with prior work~\citep{ouyang2022training, wang2023selfinstruct, chung2022scaling}, which has demonstrated that instruction tuning improves factual robustness and enhances generalization in knowledge-intensive tasks.

\textbf{Obs. 1:}  
\textbf{Factuality gaps widen with training but are consistently smaller in instruction-tuned models.}  
Table~\ref{tab:qa} reports results for models at early stopping and full convergence. We observe that the average factuality gap increases as training progresses. Across both Llama and Mistral architectures, instruction-tuned models consistently exhibit smaller gaps than their base counterparts. This pattern also holds on the WikiBios dataset.

%\textbf{Observations.} Table~\ref{tab:qa} presents evaluation results using models that reached early stopping and full convergence. We observe that the average factuality gap increases from the early stopping phase to the convergence phase. Notably, instruction-tuned models show a consistently smaller factuality gap compared to their base counterparts across both Llama and Mistral architectures. A similar factuality gap can also be observed on the WikiBios dataset.

\subsection{Factuality Gap under Out-of-distribution Generalization}

\textbf{Settings.}
Beyond in-distribution (ID) generalization, we extend factuality generalization into other two out-of-distribution (OOD) types based on the distance between the test and training data patterns: (1) \textit{near in-distribution generalization} and (2) \textit{open-world model factuality}. In the following, we examine the effects of unknown data on each type of factuality. We employ all-MiniLM-L6-v2\footnote{\url{https://huggingface.co/sentence-transformers/all-MiniLM-L6-v2}} embedding model~\cite{reimers-2019-sentence-bert} to extract and process data patterns from both OOD and ID test sets. By comparing the cosine similarity between these patterns, we are able to measure the distance between OOD and ID data. 

We conduct validation experiments using models fine-tuned on the Entity Questions dataset in Section~\ref{sec:racap1}.  For near in-distribution tasks, we sample non-overlapping data from the Entity Questions and PopQA datasets to create near in-distribution test sets, \texttt{eq\_ood} and \texttt{pop\_ood}. For the open-world task, we choose MMLU to create a complete \texttt{mmlu\_ood} set, which provides more diverse data and significantly different question formats. The cosine similarities between \texttt{eq\_ood}, \texttt{pop\_ood}, \texttt{mmlu\_ood} and the ID test set are 0.86, 0.82 and 0.55 respectively.  More details about experiments can be found in Appendix~\ref{app:ood}.

\textbf{Obs. 2:}  
\textbf{Factuality gaps persist on the OOD data but vanish under strong distribution shifts.}  
As shown in Table~\ref{tab:genralization}, Llama3.1-8B fine-tuned on known data consistently outperforms its unknown-trained counterpart on both \texttt{eq\_ood} and \texttt{pop\_ood}, with gaps of 9\% and 4\% at early stopping, and 7.5\% and 8\% at convergence. Similar trends hold for Mistral. However, on the \texttt{mmlu\_ood} dataset, which is more semantically distant, the factuality gap nearly disappears across all models.

 %\textbf{Observations.} As shown in Table~\ref{tab:genralization}, Llama3.1-8B fine-tuned on known data consistently outperforms the model fine-tuned on unknown data for both \texttt{eq\_ood} and \texttt{pop\_ood} datasets. The performance gap is 9\% on \texttt{eq\_ood} and 4\% on \texttt{pop\_ood} with early stopping, and 7.5\% and 8\% at convergence, respectively. This phenomenon can also be observed on Mistral. On \texttt{mmlu\_ood} dataset, the factuality gap even nearly disappears across all models.

\begin{table*}[t]
\begin{tabular}{ccccccccccc}
\hline
\multicolumn{2}{c}{\multirow{2}{*}{Dataset}}             & \multirow{2}{*}{Split} & \multicolumn{2}{c}{Llama} & \multicolumn{2}{c}{Llama-Instruct} & \multicolumn{2}{c}{Mistral} & \multicolumn{2}{c}{Mistral-Instruct} \\ \cline{4-11} 
\multicolumn{2}{c}{}                                     &                        & ES          & Con.        & ES               & Con.            & ES           & Con.         & ES                & Con.             \\ \hline
\multirow{2}{*}{ID}         & \multirow{2}{*}{eq\_id}    & Unknow                 & 28.25       & 24.80       & 28.75            & 25.00           & 21.15        & 18.00        & 26.00             & 20.90            \\
                            &                            & Known                  & 40.30       & 38.50       & 39.20            & 37.70           & 36.05        & 34.45        & 35.40             & 34.50            \\ \hline
\multirow{4}{*}{NID}    & \multirow{2}{*}{eq\_ood}   & Unknown                & 30.00       & 28.93       & 31.67            & 30.43           & 32.17        & 23.73        & 30.43             & 24.13            \\
                            &                            & Known                  & 39.03       & 36.60       & 38.17            & 37.03           & 34.83        & 33.00        & 34.17             & 32.43            \\ \cline{2-11} 
                            & \multirow{2}{*}{pop\_ood}  & Unknown                & 28.17       & 23.79       & 19.00            & 19.42           & 23.13        & 20.19        & 25.89             & 22.74            \\
                            &                            & Known                  & 32.58       & 32.05       & 27.54            & 25.47           & 28.69        & 27.40        & 29.71             & 28.06            \\ \hline
\multirow{2}{*}{OW} & \multirow{2}{*}{mmlu\_ood} & Unknown                & 66.11       & 66.70       & 69.23            & 69.30           & 62.63        & 62.46        & 62.25             & 62.53            \\
                            &                            & Known                  & 67.05       & 67.09       & 69.51            & 69.47           & 62.98        & 63.54        & 60.74             & 60.70            \\ \hline
\end{tabular}
\vspace{-0.2cm}
\caption{Generalization factuality. ID: in-distribution, NID: near in-distribution, OW: open world.}
\label{tab:genralization}
\vspace{-0.2cm}
\end{table*}

\subsection{A Graph-Theoretic Understanding of Factuality Gap}
\noindent\textbf{Theoretical Insight.}  
We present a formal graph-theoretic framework for analyzing factuality in LLMs. Prior work has explored knowledge extraction empirically using graphs~\citep{DBLP:conf/sigir/Tang00SSCY024,liu-etal-2024-knowledge-graph}, but lacks a principled account of generalization. We show that fine-tuning induces an edge-completion process over a latent knowledge graph, where one-hop connectivity captures factual prediction. This explains why known knowledge enables stronger generalization and why the factuality gap vanishes under semantic shift. Our analysis provides the first theoretical explanation of factuality emergence and decay in LLMs.

%Many research works~\citep{DBLP:conf/sigir/Tang00SSCY024,liu-etal-2024-knowledge-graph} have explored knowledge extraction in LLM based on graph theory. Following this line of work, we argue that memorizing knowledge in LLMs can be interpreted as a process of forming connections between different entities. Therefore, this connection-building process can also be studied through the lens of graph-based analysis. 

 Let \( \mathcal{G}_r = (\mathcal{V}, \mathcal{E}_r, \mathcal{E}^{\text{sim}}) \) be a directed graph defined under a specific relation \( r \). The node set \( \mathcal{V} = \{t_1, t_2, \ldots, t_{|\mathcal{V}|}\} \) consists of entity tokens drawn from the LLM’s token space. The edge set
$\mathcal{E}_r = \{ (v_s, v_a) \in \mathcal{V}^2 \mid f_{[:,-1]}(s, r) = a \}$
captures explicit relational knowledge: an edge from \( v_s \) to \( v_a \) exists if the model, when given the input token sequence \( (s, r) \), predicts \( a \) as the next token. The similarity edge set
$\mathcal{E}^{\text{sim}}_t = \{ (v_t, v_{t'}) \in \{t\} \times \mathcal{V} \mid t' \neq t \; \text{and} \; \|t - t'\|_2\le\epsilon \}$
represents implicit connections based on embedding similarity: an edge from \( t \) to \( t' \) exists if the distance between their embeddings is less than \( \epsilon \). A more detailed definition is provided in Appendix~\ref{app:notation}.

%Let \( \mathcal{G}_r = (\mathcal{V}, \mathcal{E}_r,\mathcal{E}^{\text{sim}}) \) denote a directed graph under relation-$r$ domain, where \( \mathcal{V} = \{t_1,t_2,...,t_{|\mathcal{V}|}\} \) represents the set of entity tokens of in the LLM's token space, \( \mathcal{E}_r=\{ (v_s,v_a) \in \mathcal{V}^2 \mid f_{[:,-1]}(s,r)=a \), which means an ordered pair $(v_s,v_a)$ appears as an explicit edge precisely when, upon receiving $(s,r)$ as its input token sequence, the model produces $a$ as the next token, and  \( \mathcal{E}_t^{\text{sim}}=\{ (v_t,v_{t'}) \in \{t\} \times \mathcal{V}\mid t' \neq t \; \text{and} \;  \|t - t'\|_2\le \varepsilon  \), which means there is an implicit edge between \( t \) and \( t' \) if their distance in the embedding space is less than \( \epsilon \). A more detailed definition is shown in Appendix~\ref{app:notation}.

% \begin{lemma}[One-Hop Knowledge as a Graph Edge]
%     \label{lemma:1}
%    For a knowledge triple \( k=(s,r,a) \), \( k \) can be directly answered by LLM \emph{if and only if} \( \exists (v_s,v_a)\in\mathcal{V}\; \text{s.t.}\; v_s=s,v_a=a,\;  (v_s,v_a)\in\mathcal{E}_r.\)
% \end{lemma}

% \textbf{Remark:} This is an  definition of knowledge that can be directly answered without other context under the graph-based framework.

\textbf{Memorizing Knowledge as an Edge-Completion Process. }
Under the graph-theoretic formulation, SFT can be viewed as an edge-completion process through which the LLM acquires new \textbf{one-hop} knowledge. Formally, this corresponds to augmenting the internal knowledge graph by adding edges that connect previously disjoint or weakly connected subgraphs, thereby encoding new relational facts into the model.

\begin{lemma}[Memorizing Knowledge as an Edge-Completion Process]
\label{lemma:2}
Let \( \mathcal{D}_r \) be the training dataset for relation \( r \). For a knowledge triple \( k = (s, r, a) \in  \mathcal{D}_{r} \), let \( \mathcal{G}_s = (\mathcal{V}_s, \mathcal{E}_s^{\text{sim}}) \) and \( \mathcal{G}_a = (\mathcal{V}_a, \mathcal{E}_a^{\text{sim}}) \) be the subgraphs connected to \( s \) and \( a \) via similarity edges. Then after memorizing \( k \), the relation graph \( \mathcal{G}_r \) is updated as  
\( \mathcal{G}_r \leftarrow \mathcal{G}_r \cup \{ (v_i, v_j) \mid (v_i, v_j) \in \mathcal{V}_s \times \mathcal{V}_a,\, f_{[:,-1]}(i, r) = j \} \).
\end{lemma}

\begin{remark}
We interpret the memorization of knowledge in LLMs as an edge-completion process on the relation graph. The formal justification is provided in Appendix~\ref{app:lemma2}. Notably, while the update considers all candidate pairs in \( \mathcal{V}_s \times \mathcal{V}_a \), only a subset of edges,specifically those satisfying \( f_{[:,-1]}(i, r) = j \), are actually added. In particular, there always exists at least one edge \( (v_s, v_a) \) added to the graph.
\end{remark}

\noindent The generalization capability of SFT is reflected in the emergence of new connections between previously unlinked subgraphs. Let \( \mathcal{G}_r' = (\mathcal{V}, \mathcal{E}_r', \mathcal{E}^{\text{sim}}) \) denote the relation graph internal to the LLM after fine-tuning. If there exists a pair \( (s', a') \in \mathcal{V}_s \times \mathcal{V}_a \) such that the corresponding knowledge triple \( (s', r, a') \notin \mathcal{D}_r \), and  
\( (v_{s'}, v_{a'}) \notin \mathcal{E}_r \) but \( (v_{s'}, v_{a'}) \in \mathcal{E}_r' \),  
then the model has successfully generalized beyond the training data by inferring the unseen triple \( (s', r, a') \).

\begin{figure}[t]
    \centering
    \includegraphics[width=\linewidth]{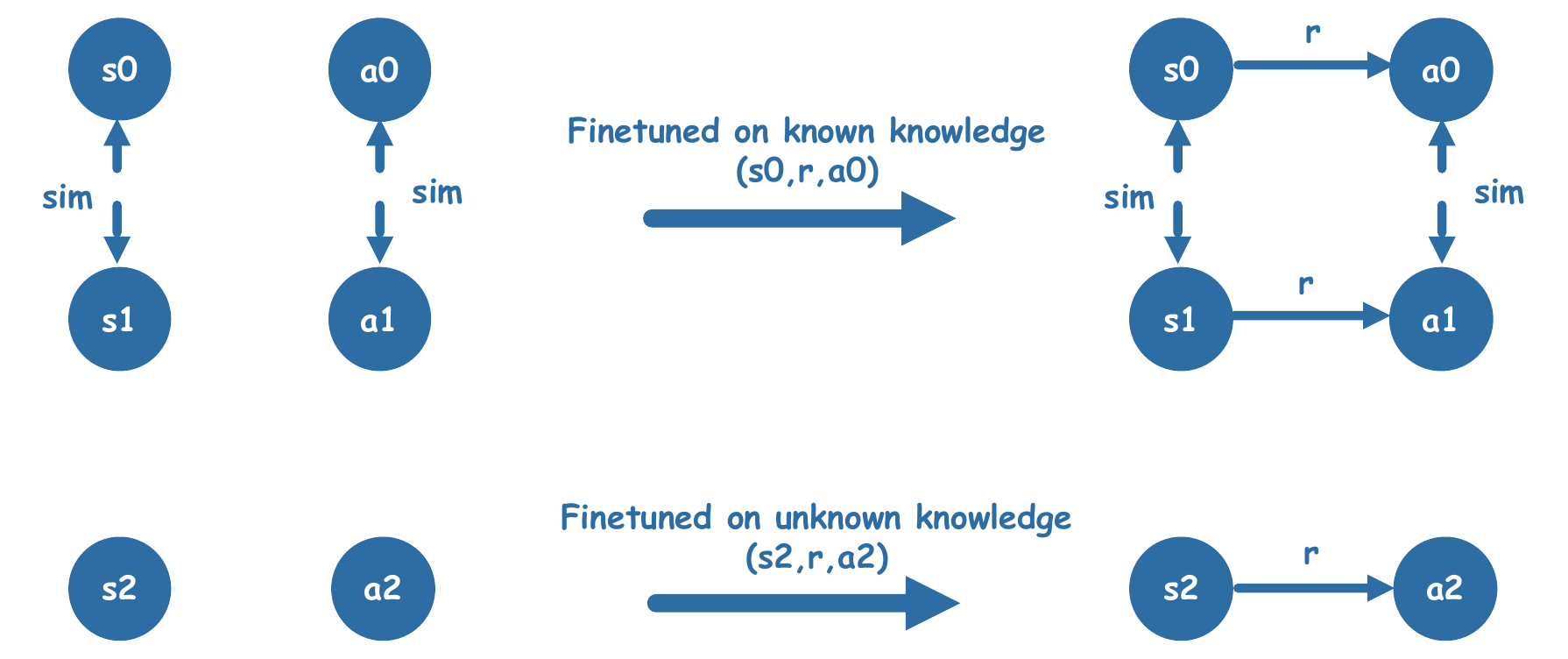}
    \caption{Memorizing a known knowledge triple $(s_0, r, a_0)$ generalizes to memorizing $(s_1, r, a_1)$ but memorizing an unknown knowledge triple $(s_2, r, a_2)$ can not generalize.}
    \label{fig:sft_graph}
    \vspace{-0.2cm}
\end{figure}

\textbf{Factuality Gap Explained via Differential Connectivity. }
If a knowledge triple \( (s, r, a) \) is present in the training set \( \mathcal{D}_r \), few-shot prompting can be viewed as temporarily injecting edges \( \{(v_{s'_i}, v_{a'_i})\} \), where each \( (s'_i, r, a'_i) \) is a support triple, to connect \( \mathcal{V}_s \) and \( \mathcal{V}_a \). This mechanism will be presented  in Section~\ref{sec:role_of_icl}.
Unknown knowledge under few-shot prompting typically arises when the connectivity between \( v_s \) and \( v_{s'_i} \), or between \( v_a \) and \( v_{a'_i} \), is weak, which is often due to sparsity in the induced subgraphs, particularly when \( s \) or \( a \) corresponds to a low-degree entity. As a result, fine-tuning on such unknown knowledge induces only limited updates to the relation graph, thereby reducing the model’s capacity to generalize across the domain \( r \).
Figure~\ref{fig:sft_graph} illustrates the process of adding edges when LLM finetuned on different types of knowledge. 

\begin{theorem}[Factuality Gap as a Connectivity Gap in Knowledge Graphs]
\label{theorem:gap}
Let \( \mathcal{G}_{\text{kn}} = (\mathcal{V}, \mathcal{E}_{\text{kn}}, \mathcal{E}^{\text{sim}}) \) and \( \mathcal{G}_{\text{unk}} = (\mathcal{V}, \mathcal{E}_{\text{unk}}, \mathcal{E}^{\text{sim}}) \) be knowledge graphs induced by LLMs fine-tuned on known and unknown knowledge, respectively. Let \((s, r, a)\) be a test triple sampled uniformly at random from a fixed test set. Define indicator variables \( Z_{\text{kn}} = \mathbf{1}\{(v_s, v_a) \in \mathcal{E}_{\text{kn}}\} \) and \( Z_{\text{unk}} = \mathbf{1}\{(v_s, v_a) \in \mathcal{E}_{\text{unk}}\} \). Assume edges under relation \(r\) are uniformly distributed and test triples are uniformly sampled over their support. Then the expected factuality gap satisfies
$\mathbb{E}[Z_{\text{kn}} - Z_{\text{unk}}] = \Delta_{\text{fact}} \propto |\mathcal{E}_{\text{kn}}| - |\mathcal{E}_{\text{unk}}| > 0.$
\end{theorem}
\begin{remark}
Theorem~\ref{theorem:gap} interprets the factuality gap as a direct consequence of differences in one-hop connectivity induced by fine-tuning. Under uniform sampling assumptions, the expected gap in one-hop accuracy reflects the difference in the number of factual edges established in the graph. A detailed proof is provided in Appendix~\ref{app:proof_gap}.
\end{remark}
\textbf{OOD Generalization and the Vanishing Gap. }
As the test distribution diverges from the training graph structure, both known and unknown knowledge graph $\mathcal{G}_{\text{kn}}, \mathcal{G}_{\text{unk}}$ become equally less overlapped to the OOD knowledge graph. Consequently, as the knowledge graph corresponding to the training data domain becomes nearly disjoint from that of the test data domain, the factuality gap approaches zero.

\begin{table*}[t!]
\setlength{\tabcolsep}{2pt}  % 设置列间距
\centering
\begin{tabular}{cc|cccccccc}
\hline
\multicolumn{2}{c}{\multirow{2}{*}{Dataset}} & \multicolumn{2}{c}{Llama} & \multicolumn{2}{c}{Llama-Instruct} & \multicolumn{2}{c}{Mistral} & \multicolumn{2}{c}{Mistral-Instruct} \\
\cline{3-10}
\multicolumn{2}{c}{} & ES & Con. & ES & Con. & ES & Con. & ES & Con. \\
\hline

\multirow{2}{*}{\rotatebox[origin=c]{90}{EQ}} 
& U & \rule{0pt}{10pt} 41.55$\scriptstyle{+}13.3$ & 38.95$\scriptstyle{+}14.2$ & 41.00$\scriptstyle{+}12.3$ & 37.40$\scriptstyle{+}12.4$ & 35.35$\scriptstyle{+}14.2$ & 32.95$\scriptstyle{+}15.0$ & 35.25$\scriptstyle{+}9.25$ & 30.05$\scriptstyle{+}9.15$ \\
& K & \rule{0pt}{10pt} 43.45$\scriptstyle{+}3.15$ & 42.20$\scriptstyle{+}3.70$ & 41.20$\scriptstyle{+}2.00$ & 40.70$\scriptstyle{+}3.00$ & 38.25$\scriptstyle{+}2.20$ & 37.95$\scriptstyle{+}3.50$ & 33.15$\scriptstyle{-}2.25$ & 32.65$\scriptstyle{-}1.85$ \\
\hline

\multirow{2}{*}{\rotatebox[origin=c]{90}{PQ}} 
& U & \rule{0pt}{10pt} 39.82$\scriptstyle{+}8.54$ & 37.89$\scriptstyle{+}10.91$ & 35.06$\scriptstyle{+}4.97$ & 34.01$\scriptstyle{+}6.68$ & 35.93$\scriptstyle{+}8.99$ & 35.76$\scriptstyle{+}15.22$ & 31.46$\scriptstyle{+}6.20$ & 31.32$\scriptstyle{+}11.73$ \\
& K & \rule{0pt}{10pt} 38.77$\scriptstyle{+}1.96$ & 38.66$\scriptstyle{+}3.11$ & 35.55$\scriptstyle{-}0.31$ & 36.18$\scriptstyle{+}1.09$ & 35.93$\scriptstyle{+}2.93$ & 35.90$\scriptstyle{+}4.23$ & 31.63$\scriptstyle{-}0.63$ & 31.84$\scriptstyle{+}0.21$ \\
\hline

\multirow{2}{*}{\rotatebox[origin=c]{90}{MU}} 
& U & \rule{0pt}{10pt} \underline{54.80$\scriptstyle{+}19.9$} & \underline{54.60$\scriptstyle{+}20.7$} & \underline{64.99$\scriptstyle{+}31.4$} & \underline{65.32$\scriptstyle{+}31.8$} & \underline{55.39$\scriptstyle{+}27.3$} & \underline{55.13$\scriptstyle{+}28.6$} & 58.00$\scriptstyle{+}26.4$ & 60.09$\scriptstyle{+}34.2$ \\
& K & \rule{0pt}{10pt} \underline{67.60$\scriptstyle{+}30.1$} & \underline{67.86$\scriptstyle{+}30.8$} & \underline{69.30$\scriptstyle{+}33.40$} & \underline{68.84$\scriptstyle{+}34.0$} & \underline{58.46$\scriptstyle{+}22.9$} & \underline{58.39$\scriptstyle{+}23.6$} & 61.07$\scriptstyle{+}27.6$ & 60.94$\scriptstyle{+}28.8$ \\
\hline

\multirow{2}{*}{\rotatebox[origin=c]{90}{WB}} 
& U & \rule{0pt}{10pt} \underline{55.20$\scriptstyle{-}0.30$} & \underline{48.32$\scriptstyle{+}1.42$} & & & \underline{47.93$\scriptstyle{+}0.63$} & \underline{37.99$\scriptstyle{+}1.32$} & & \\
& K & \rule{0pt}{10pt} \underline{58.20$\scriptstyle{-}0.05$} & \underline{50.85$\scriptstyle{+}1.16$} & & & \underline{50.58$\scriptstyle{+}1.42$} & \underline{40.22$\scriptstyle{+}0.64$} & & \\
\hline
\end{tabular}
\vspace{-0.2cm}
\caption{Performance of the fine-tuned model with few-shot and few-shot CoT. EQ: Entity Questions, PQ: PopQA, MU: MMLU, WB: WikiBios.  Exact Match Accuracy for QA tasks and FactScore for WikiBios, with \underline{underlined} results for few-shot and non-underlined for few-shot CoT. The small number in the bottom right corner represents the improvement or decline in current performance relative to the performance without using few-shot learning.}
\vspace{-0.2cm}
\label{tab:cot}
\end{table*}

 \begin{theorem}[Decay of Factuality Gap Under Distributional Shift]
\label{theorem:ood}
Let  $\cos\langle \mathcal{D}_{\text{test}}, \mathcal{D}_{\text{train}} \rangle := \mathbb{E}_{x \sim \mathcal{D}_{\text{test}},\, x' \sim \mathcal{D}_{\text{train}}} \langle x, x' \rangle$
denote the semantic similarity between the test and training distributions, where \(x, x'\) are unit-normalized representations. Then, the factuality gap \(\Delta_{\text{fact}}\) under OOD evaluation decreases as the semantic similarity vanishes: if $\cos\langle \mathcal{D}_{\text{test}}, \mathcal{D}_{\text{train}} \rangle \to 0$, the factuality gap $\Delta_{\text{fact}} \to 0.$

\end{theorem}
\begin{remark}
In practice, we compute the semantic similarity using an external embedding model. We assume that the resulting scores closely approximate those that would be obtained using the internal representations of the LLM. A formal proof of Theorem~\ref{theorem:ood} is provided in Appendix~\ref{app:proof_ood}.
\end{remark}

% \begin{tcolorbox}[colback=gray!5, colframe=gray!80, title=Theoretical Takeaway, fonttitle=\bfseries\color{black}]
% Fine-tuning on known knowledge completes one-hop edges in the model’s graph, enabling generalization. Unknown knowledge leads to sparse updates and limited reach. As the test data shift away, this structural gap shrinks—explaining the decay of the factuality gap.

% \end{tcolorbox}

\section{Can Fatcuality Gap be Easily Mitigated? (RQ2)} \label{sec:ood}

\subsection{ICL Mitigates the Factuality Gap}
\textbf{Settings.}  
We evaluate all models and tasks from Section~\ref{sec:racap} using few-shot and few-shot CoT prompting. Few-shot examples are selected from the \textit{Known} training data. For CoT, GPT-4o\footnote{\url{https://openai.com/index/gpt-4o-system-card/}} generates entity-level analyses to construct reasoning chains, which are integrated into the CoT prompts. The format is shown below.

\begin{tcolorbox}[
    colframe=black,
    colback=gray!15,
    coltitle=black,
    fonttitle=\bfseries,
    enhanced,
    rounded corners,
    left=6pt, right=6pt, top=6pt, bottom=6pt,
    boxrule=1pt,
    arc=6pt,
    width=\linewidth
]
\textit{Question:\{\} Analysis:\{\} Answer:\{\}}
\label{box:cot}
\end{tcolorbox}
\noindent We construct three prompt sets and evaluate two prompting variants: with and without CoT. All models, including \textit{Known} and \textit{Unknown}, are evaluated using the same prompts. The prompt set yielding the highest performance on the \textit{Unknown} model is reported. For generation tasks, we use few-shot prompting only, following the same example selection strategy. Full prompt details are provided in Appendix~\ref{app:cot}.

\textbf{Obs. 3:}  
\textbf{In-context learning narrows the factuality gap, especially with few-shot CoT.}  
% Table~\ref{tab:cot} compares model performance under few-shot and few-shot CoT prompting across different datasets. We observe that few-shot prompting improves performance more significantly on the \textit{Unknown} split than on the \textit{Known} split, indicating that the factuality gap can be reduced—or even eliminated—via prompting. We highlight three key findings:
% (1) The gap is more easily mitigated in models trained with early stopping.  
% (2) \textit{Instruct} models exhibit greater gap reduction than \textit{Base} models, particularly at convergence.  
% (3) On MMLU and WikiBios, few-shot prompting sometimes increases the gap. We attribute this to task complexity: MMLU features highly diverse question formats, while WikiBios involves open-ended generation, both of which lead to more entangled factuality patterns.
Table \ref{tab:cot} presents a comparison of the results obtained through few-shot or few-shot CoT inference after training different models on various datasets. We can observe that, in most cases, after using few-shot learning, the performance on the \textit{Unknown} split improves more significantly compared to the \textit{Known} split. This suggests that the factuality gap can be mitigated or even fully eliminated. Additionally, we observe the following points: 1) The gap in models with early stopping is more easily mitigated.  2) The factuality gap of the \textit{Instruct} model is easier to mitigate than \textit{Base} model, especially in the case of Convergence. In MMLU and WikiBios, using few-shot learning sometimes even increases the performance gap. This may be due to the particularities of these two tasks compared to regular QA tasks. The former is a comprehensive dataset with complex and varied question formats, while the latter is an open-ended generation task, both of which result in a more complex factuality gap pattern.

\subsection{Ablation study}
To better understand how in-context learning mitigates the factuality gap, we conduct ablation experiments using the Llama-3.1-8B model on the Entity Questions dataset. Full details are provided in Appendix~\ref{app:aba}.

\begin{figure}[t]
\centering
 \includegraphics[width=0.9\linewidth]{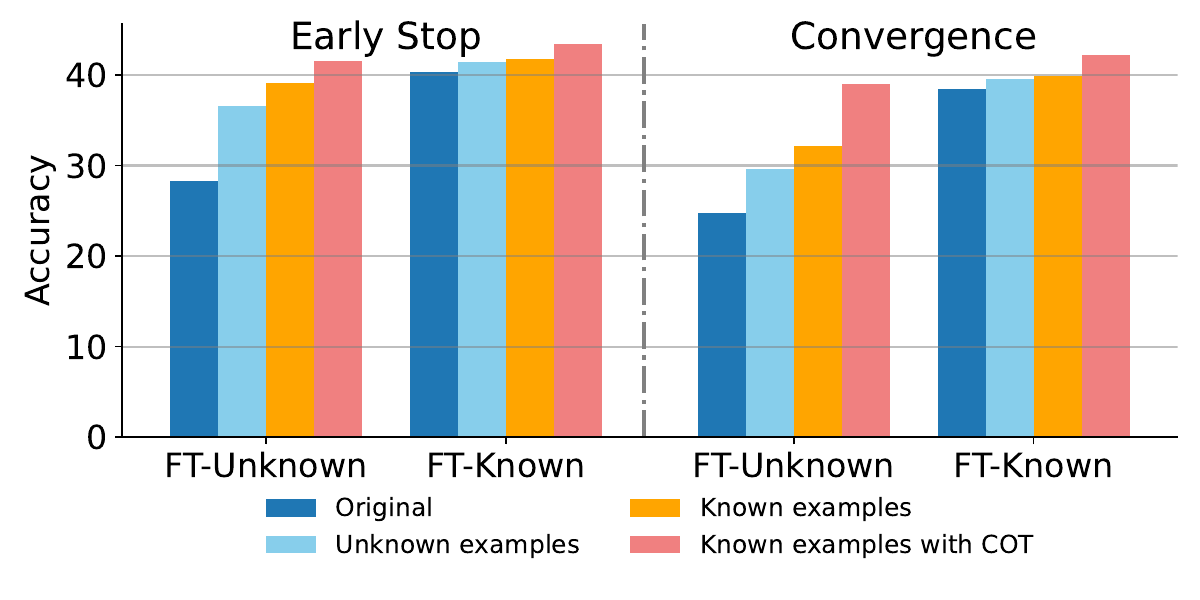}
  \vspace{-0.2cm}
  \caption{Ablation study of few-shot examples and CoT.}
  \label{fig:fewshot}
  \vspace{-0.2cm}
\end{figure}

\begin{figure}[t]
\centering
\includegraphics[width=0.9\linewidth]{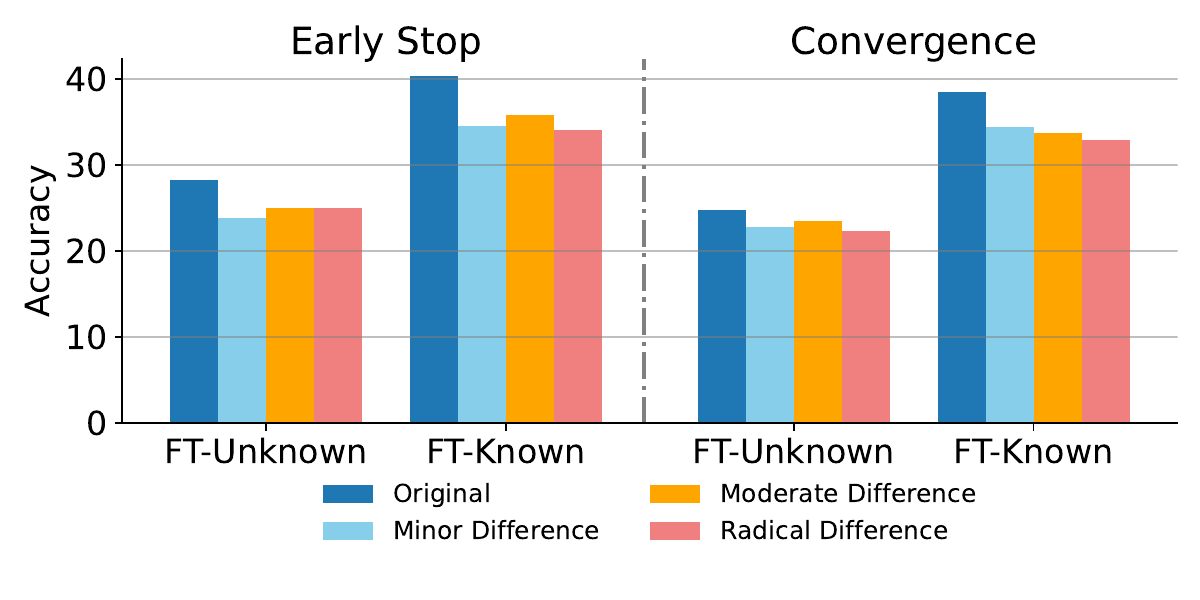}
  \vspace{-0.2cm}
  \caption{Ablation study of prompt formulation. We use three levels of rephrasing: Minor, Moderate, Radical.}
  \vspace{-0.2cm}
  \label{fig:rephrase}
\end{figure}

\textbf{Obs. 4: All of example source, CoT reasoning and question format critically affect factual generalization.}  
We conduct an ablation study on the composition of the prompt, separately examining the source of examples in few-shot prompts and the impact of CoT. We validated the effectiveness of \textit{Known} examples and CoT, as shown in Figure \ref{fig:fewshot}. We also study the impact of changing the prompt format on the factuality gap. We use GPT-4o to rephrase these questions in three different formats and find that the performance decline in all cases, and the factuality gap remains large, which is illustrated in Figure~\ref{fig:rephrase}.

\iffalse To better understand the essence of how in-context learning mitigates the factuality gap, we design the following ablation experiment on Llama Base model and Entity Questions dataset. Details of ablation studies can be found in Appendix \ref{app:aba}.

\textbf{Prompt components and formulation. }
%fewshot cot
We conduct an ablation study on the composition of the prompt, separately examining the selection of examples in few-shot prompts and the impact of CoT. We validated the effectiveness of \textit{Known} examples and CoT, as shown in Figure \ref{fig:fewshot}. 
We also study the impact of changing the prompt format on the factuality gap. We use GPT-4o to rephrase these questions in three different formats and find that the performance decline in all cases, and the factuality gap remains large.\fi

\subsection{Understanding the Role of ICL}
\label{sec:role_of_icl}
\textbf{ICL prompts act as subgraph injections that reduce the factuality gap.}  
We present a new theoretical perspective on ICL: few-shot examples and CoT rationales can be interpreted as \textit{prompt-induced subgraphs} that augment the model's internal knowledge graph during inference.
Given a prompt \( \Pi \) containing \( n \) support triples \( (s_i, r, a_i) \), we treat it as an auxiliary knowledge graph \( \mathcal{G}_\Pi = (\mathcal{V}_\Pi, \mathcal{E}_{r_\Pi}, \mathcal{E}^{\text{sim}}_\Pi) \). For CoT prompting, a target triple \( (s, r, a) \) is supported by a structured reasoning chain  
$C = \{(s, r_i, a_i) \mid 1 \leq i \leq n, \; a_n = a\},$
where all steps share the subject \( s \), and relations \( r_i \) may differ. This defines an additional support graph \( \mathcal{G}_C \).
At inference time, the model operates on an augmented graph
$\mathcal{G}^\star = \mathcal{G} \cup \mathcal{G}_\Pi \cup \mathcal{G}_C,$
where \( \mathcal{G} \) is the base knowledge graph encoded by the fine-tuned model, and \( \mathcal{G}_\Pi \cup \mathcal{G}_C \) are injected through the ICL prompt.

\iffalse
\textbf{Few-shot Examples and CoT Adds a Prompt Subgraph. }
In few-shot prompting, given a prompt \( \Pi \) consisting of \( n \) explicit triples \( (s_i, r, a_i) \) and their corresponding token representations, we treat \( \Pi \) as a knowledge graph \( \mathcal{G}_\Pi = (\mathcal{V}_\Pi, \mathcal{E}_{r_{\Pi}}, \mathcal{E}^{\text{sim}}_{\Pi})\)  in LLM. Figure~\ref{fig:icl_graph} illustrates the comparison of added edges resulting from ICL on both the FT-Known and FT-Unknown models.
In CoT prompting, a knowledge triple \( (s, r, a) \) is supported by a structured sequence of $n$ auxiliary triples:
\(
C=\{(s, r_i, a_i)\mid 1 \leq i \leq n, a_n=a\},
\)
where all triples share the same subject token \( s \), but the relations \( r_i \) may vary. We treat CoT prompt as a graph \( \mathcal{G}_C \) in LLM. At inference time,  we construct an augmented graph \( \mathcal{G^\star}=\mathcal{G} \cup \mathcal{G}_{\Pi} \cup \mathcal{G}_C\) with \( \Pi\) and  \(C\) prompted into the LLM.  
\fi

\begin{theorem}[ICL Prompt Can Mitigate the Factuality Gap]
\label{theorem:icl}
Let \( \mathcal{G} \) be the knowledge graph induced by an LLM after fine-tuning, and let \( \mathcal{P} \) be a valid in-context prompt represented as an auxiliary graph \( \mathcal{G}_{\mathcal{P}} \). The augmented graph at inference time is
$\mathcal{G}^\star = \mathcal{G} \cup \mathcal{G}_{\mathcal{P}}.$
Then, the factuality gap under prompt-augmented inference satisfies
$\Delta^\star_{\text{fact}} < \Delta_{\text{fact}}.$
\end{theorem}
\begin{remark}
This result provides a structural explanation for why in-context prompting improves factuality: it temporarily densifies connectivity between relevant subgraphs, effectively compensating for missing fine-tuned edges. Please refer to Appendix~\ref{app:proof_icl} for the proof of Theorem~\ref{theorem:icl}.

% To our knowledge, this is the first formal interpretation of ICL as prompt-induced graph completion, and the first proof that such augmentation strictly reduces the factuality gap.
\end{remark} 
\begin{figure}[t]
\centering
  \vspace{-0.2cm}
  \includegraphics[width=\linewidth]{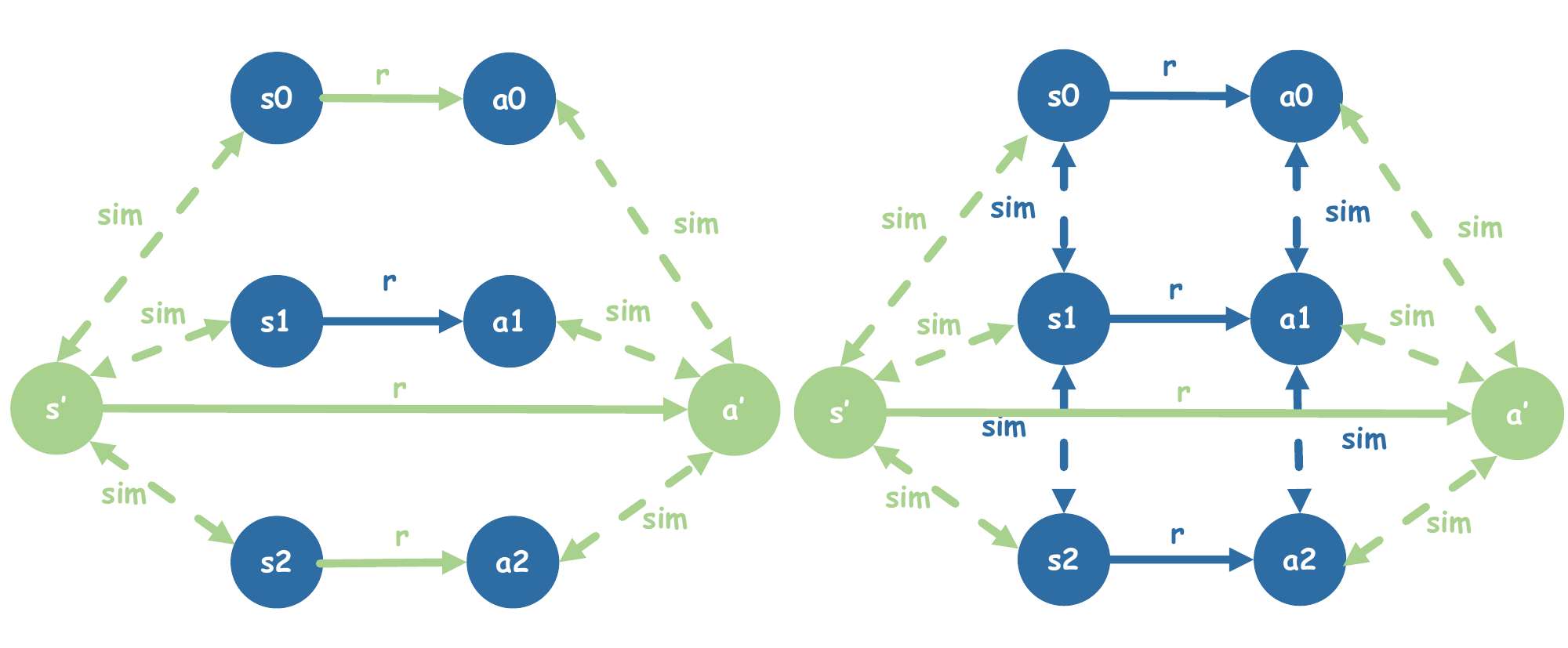}
    \vspace{-0.2cm}
  \caption{In an LLM fine-tuned on unknown knowledge (left), the demonstration $(s', r, a')$ introduces new edges $(s_0, a_0)$ and $(s_2, a_2)$. In contrast, for the LLM fine-tuned on known knowledge (right), these edges already exist and thus are not newly added. Consequently, the factuality gap narrows as the difference in the number of edges between the two graphs decreases.}
  \label{fig:icl_graph}
  \vspace{-0.2cm}
\end{figure}

\section{Leveraging ICL for Knowledge Extraction (RQ3)}

% \textbf{Identifying High-Quality Fine-Tuning Data.}  
% Few-shot prompting has been used to identify high-quality fine-tuning data~\citep{gekhman-etal-2024-fine}. Our work validates the effectiveness of this approach both empirically and theoretically.
% In the last section, we theoretically demonstrated that in-context learning (ICL) helps reduce the factuality gap. In this section, we complement that analysis with empirical evidence, using the same experimental framework to validate the practical effectiveness of the ICL approach.

%数据质量不好或者数据量不足的时候使用ICL比不用ICL效果/不训练 更好
\subsection{Improving Generalization under Limited or Noisy Supervision}  
\begin{figure}[h]
\centering
  \includegraphics[width=\linewidth]{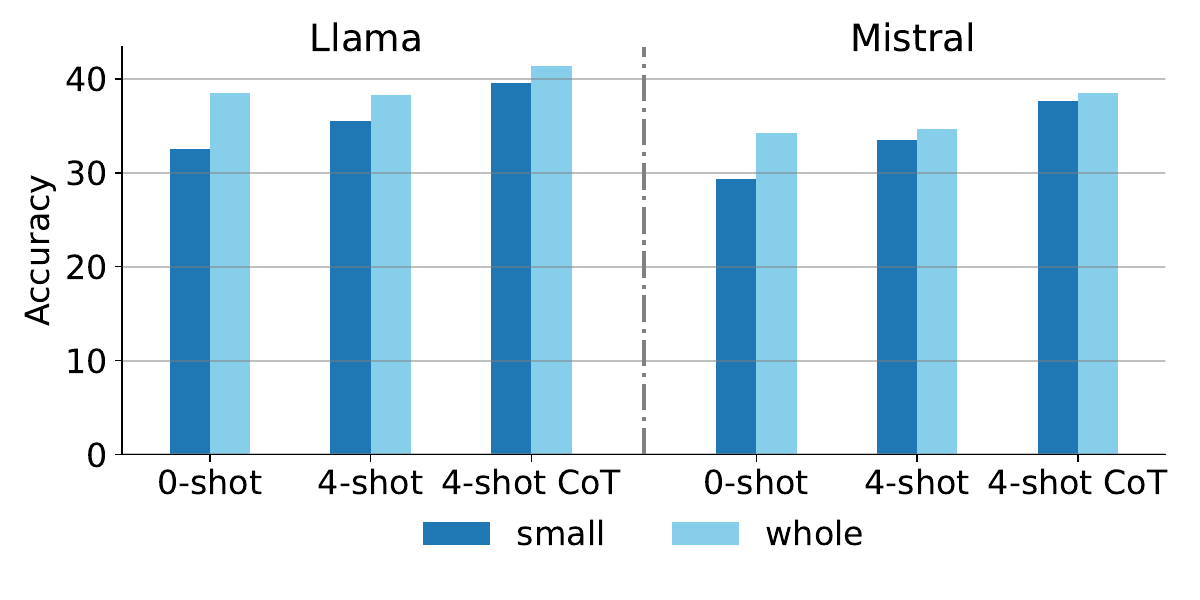}
  \caption{Comparison between Llama-3.1-8B and Mistral-7B-v0.3 models fine-tuned on 5\% of dataset and the whole dataset.}
  \vspace{-0.2cm}
  \label{fig:small}
    \vspace{-0.2cm}
\end{figure}

Building on our theoretical insights, we hypothesize that ICL can improve factual generalization not only in the presence of low-quality fine-tuning data, but also when the available data is limited. To test this, we conduct an experiment on the PopQA dataset, comparing two conditions: (1) applying ICL after fine-tuning on a random 5\% subset of the training data, and (2) applying ICL after full-data fine-tuning. As shown in Figure~\ref{fig:small}, the 5\%-trained model achieves performance comparable to the full-data model when combined with ICL. Full details are provided in Appendix~\ref{app:small}. These findings suggest that well-designed ICL prompts can effectively compensate for limited or low-quality supervision in the knowledge extraction of LLM.

\subsection{Rethinking the Metric for Finetuning Data Selection Method}
% 数据选择方法，在evaluation上用few-shot来测试是可能是不公平的
% Few-shot prompting has been used to identify high-quality fine-tuning data~\citep{gekhman-etal-2024-fine}. 
% Many recent studies on fine-tuning data selection algorithms evaluate their final performance using few-shot prompting~\citep{liu2024tsds,xia2024less}. However, based on the theoretical and experimental results presented in this paper, we argue that ICL can effectively reduce, and in some cases even eliminate, the performance gap caused by the differences in training data. As a result, evaluation based solely on few-shot prompting may obscure the true effectiveness of data selection algorithms. A more comprehensive evaluation protocol is needed to reliably assess their impact.
Recent studies on data selection algorithms for fine-tuning commonly evaluate performance using few-shot prompting~\citep{liu2024tsds, xia2024less}. However, our theoretical and empirical findings suggest that in-context learning can significantly reduce, and in some cases even eliminate, the performance differences arising from variations in training data. Consequently, evaluations based solely on few-shot prompting may mask the true effectiveness of data selection methods. We therefore argue that a more comprehensive evaluation framework is necessary to reliably assess the performance of data selection algorithms.

% \begin{figure}[th]
% \centering
%   \includegraphics[width=0.6\linewidth]{figures/train_acc_loss_toyExample.pdf}
%   \caption{Training process of ToyExample on Entity Questions (EQ).}
%   \label{fig:train_acc_loss}
% \end{figure}
\section{Discussion: How Far Can CoT Go?}
\label{toyExampleUp}
% To isolate the effect of knowledge type on factual generalization, we design a controlled toy example that removes confounding factors such as data quality and filtering noise. 
% While earlier experiments show that ICL reduces the factuality gap, the improvement is often modest. This synthetic setup tests whether the gap can be further mitigated under idealized conditions.

\textbf{Toy Example Setup.}  
To further eliminate the potential impact of data filtering, we construct a Toy Example using manually created \textit{Unknown} data that genuinely extends beyond the knowledge boundary of the LLM. We use the Llama3.3-70B-Instruct\footnote{\url{https://huggingface.co/meta-llama/Llama-3.3-70B-Instruct}} model to extract data from the EntityQuestions dataset with a single query, without relying on few-shot examples.  We then introduce fixed-format perturbations ("\$\&") to entity tokens in the known set to create unknown knowledge set, ensuring that the model is unable to handle these perturbed examples. We fine-tune the models using LoRA, and evaluate their performance on the test set, which shares the same data type as the training set, i.e., normal (known) or perturbed (unknown). Additionally, we also add special CoT to the Toy Example for verification.  Detailed prompt design is shown in the box below. Experiment details are presented in  Appendix \ref{app:toy}.

\begin{tcolorbox}[
    title=CoT for Toy Example,
    colframe=gray, % Frame color
    colback=gray!15,  % Background color
    coltitle=gray, % Title color
    fonttitle=\bfseries\color{white}, % Bold title font
    rounded corners, % Sharp corners for the box
    enhanced, % Additional styling options
    left=6pt, right=6pt, top=6pt, bottom=6pt, % Padding
    boxrule=1pt, % Border thickness
    arc=6pt,
    width=\linewidth % Box width
]
 Ignore all the special characters in the following question. Think step by step. First, clean all special characters in the question. In this step, you might see some Unicode characters in foreign languages. Next, rethink the cleaned question. Finally, give the detailed answer of the cleaned question with a short explanation. 
\label{box:cot}
\end{tcolorbox}

\textbf{Obs 5:}  
\textbf{Under controlled perturbations, the factuality gap remains large, but is substantially reduced by CoT prompting.}  
As shown in  Table ~\ref{tab:toy13}, we observe consistent gaps in  factuality across models fine-tuned on known and unknown knowledge sets. The results further confirm that unknown knowledge encourages factuality failure.  We also observe that CoT effectively enhances model testing performance and narrows the factuality gap between the two 70B models.
% \textbf{Observations.}
% As shown in  Table ~\ref{tab:toy13}, we observe consistent gaps in  factuality across models fine-tuned on known and unknown knowledge sets. The results further confirm that unknown knowledge encourages factuality failure.  We also observe that CoT effectively enhances model testing performance and narrows the factuality gap between the two 70B models.

\begin{table}[h]
\centering
\begin{tabular}{ccccc}
\hline
\multirow{2}{*}{Split} & \multicolumn{2}{c}{Original} & \multicolumn{2}{c}{With CoT} \\ 
\cline{2-5} & ES & Con. & ES & Con.\\ \hline
Unknown& 44.73 & 41.70 & 84.08 & 82.81 \\
Known & 83.11 & 82.81 & 86.72 & 87.60 \\ \hline
\end{tabular}
\caption{Performance of Toy Example} % with and without CoT.
\vspace{-0.3cm}
\label{tab:toy13}
\end{table}

\textbf{Discussion.}
For more powerful 70B models, fine-tuning on both known and unknown knowledge can still lead to a factuality gap. However, the way these models mitigate the gap through in-context learning differs significantly from the approach discussed above. This mitigation is achieved by using instructions to directly establish a connection between perturbed entities and normal entities, which then enables correct knowledge extraction. These results demonstrate that CoT is powerful enough to bypass the mapping established during the fine-tuning stage, allowing the model to respond based on the new mapping defined within the CoT prompt. This highlights the effectiveness of prompt-based reasoning in decoupling model behavior from parameter-level modifications.

\section{Conclusion}
This work provides both theoretical and empirical investigations of the factuality gap introduced by fine-tuning LLMs on known versus unknown knowledge. Based on the analysis of experimental phenomena, we further attempt to explain and investigate this gap from a graph-theoretic perspective, viewing the process of knowledge extraction as a problem of graph connectivity and structural completeness. This theoretical framework reveals the interaction mechanism between fine-tuning and test-time ICL prompts, uncovering how prompt-based reasoning compensates for parameter-induced limitations. In summary, in this paper, we offer a new perspective on the factual behavior of LLMs, providing foundational insights into factual generalization that can inform data selection strategies, prompt design, model interpretability, and the deployment of models in knowledge-intensive tasks.

% \section{Conclusion} This work provides both theoretical and empirical investigations of the factuality gap introduced by fine-tuning LLMs on known versus unknown knowledge. Based on the analysis of experimental phenomena, we further attempt to explain and investigate this gap from a graph-theoretic perspective, viewing the process of knowledge extraction as a problem of graph connectivity. This theoretical framework reveals the interaction mechanism between fine-tuning and test-time ICL prompts. In summary, in this paper, we offer a new perspective on the LLM factuality and these novel insights offer a new foundation for understanding and improving factual generalization in LLMs, with implications for data selection, prompt design, and knowledge-intensive task deployment.

\section*{Limitations} The proposed framework is derived from empirical observations and may lack full formal generality. Some underlying assumptions may not fully capture model behavior across diverse domains, architectures, or prompt formats. In particular, this work does not fully explain the anomalous behavior observed on datasets such as MMLU and WikiBios, which may involve more complex or multimodal factual structures. We hope this work encourages future efforts to refine the theoretical framework, extend it to broader task types, and develop more robust explanations for these challenging settings.

\newpage

\bibliography{custom}

\newpage
\appendix
\section{Theory Work}
\subsection{Notation and Setup}
\label{app:notation}
% \paragraph{Embedding Layer} We consider the embedding matrix \(E\)and un-embedding matrix \(U\)modules to be weight-tied, and both of them is fixed to be the identity matrix of size \(d=|\mathcal{T}|\), where  \(\mathcal{T}\) is the token set of LLM. Under this setting, the embedding of every token is simply its corresponding standard basis vector $e_i$ (all zeros except for a 1 in the $i$-th position). And that is \(E=U=I_{|\mathcal{T}|}, \; e_i \in \mathbb{R}^d\).

\paragraph{Embedding Layer}
We define the embedding matrix \(E \in \mathbb{R}^{|\mathcal{T}| \times d}\), where the \(i\)-th row \(E[i] = E_{t_i}\) is the (non-orthogonal) embedding vector of token \(t_i\), and the un-embedding matrix is \(U \in \mathbb{R}^{d \times |\mathcal{T}|}\). The matrices \( E \) and \( U \) are weight-tied and are learned during pretraining.

We assume the embeddings are non-orthogonal and fixed during finetuning, this setting reflects realistic language model behavior and allows us to define local neighborhoods over tokens via  euclidean distance similarity in the embedding space. Specifically, we define:
\[
t_i \sim_\epsilon t_j \iff \|E[i] - E[j]\| < \epsilon,
\]
which enables generalization from seen tokens to nearby tokens in the semantic space.

% \paragraph{One-Layer Transformer Architecture}
% We consider a one-headed, one-layer transformer in this work with fully orthogonal and weight-tied embedding and un-embedding layers. We assume that the key, query, and value matrices are square projections \(W_{V},W_{KQ}\in \mathbb{R}^{d\times d}\) with \(W_{KQ}=W_{K}^TW_{Q}\), thereby preserving the dimensions of the embedding. We additionally assume that the language modeling head corresponds to an identity transformation. For a  prompt \((s,r)\), the input embedding matrix is \(X=[e_s,e_r]\in\mathbb R^{d\times2}\). Then the attention score is
% \begin{align*}
% \alpha&=\text{softmax}(X^{\top}W_{KQ}X_{:,-1})\\
% &=\text{softmax}(\left[
% \begin{array}{c}
% (W^{KQ})_{s,r} \\
% (W^{KQ})_{r,r}
% \end{array}
% \right])\\
% &=\left[\begin{array}{c}
% \alpha_{s}\\
% \alpha_{r}
% \end{array} \right].
% \end{align*}
% Then the hidden states vector is
% \begin{align*}
% h(s, r) &= W^V X \alpha \\
%         &= \alpha_s W^Ve_s + \alpha_r W^Ve_r \\
%         &= \alpha_s W^V_{[:, s]} + \alpha_r W^V_{[:, r]}.
% \end{align*}
% Because un-embedding matrix is \(U=E=I_d\) and the next token logits is
% \begin{align*}
% z_i(s, r) &= e_i^\top h(s, r) \\
%          &= \alpha_s (W^V)_{i, s} + \alpha_r (W^V)_{i, r}.
% \end{align*}
% The model probability for token \(i\) is
% \[
% p_{\theta}(i \mid s, r) = \text{softmax}_i\left(z(s, r)\right)
% \]
% We use the same one-layer transformer toy model with the work~\citep{ghosal2024understanding}, where the memory ability of this model has been proved. 

\paragraph{One-Layer Transformer Architecture}
We consider a one-headed, one-layer transformer with untied, learned embedding \(E \in \mathbb{R}^{|\mathcal{T}| \times d}\) and un-embedding \(U \in \mathbb{R}^{d \times |\mathcal{T}|}\) matrices. For a prompt \((s, r)\), the input embedding matrix is \(X = [E_s, E_r] \in \mathbb{R}^{d \times 2}\), where \(E_s, E_r\) are the continuous token embeddings.

Let \(W^{KQ} = (W^K)^\top W^Q \in \mathbb{R}^{d \times d}\) and \(W^V \in \mathbb{R}^{d \times d}\). The attention weights are:
\begin{align*}
\alpha&=\text{softmax}(X^{\top}W^{KQ}X_{:,-1})\\
&=\text{softmax}(\left[
\begin{array}{c}
(W^{KQ})_{s,r} \\
(W^{KQ})_{r,r}
\end{array}
\right])\\
&=\left[\begin{array}{c}
\alpha_{s}\\
\alpha_{r}
\end{array} \right].
\end{align*}
where the subscript $:,-1$ denotes the last column of the matrix. Thus, we take the softmax of the (post-self-attention) embedding of the last input token to predict the next token.
The hidden state is:
\[
h(s, r) = W^V X \alpha = \alpha_s W^V E_s + \alpha_r W^V E_r.
\]The logits for token \(i\) are computed via:
\[
z_i(s, r) = U_{:,i}^\top h(s, r),
\] and the output distribution is:
\[
p_\theta(i \mid s, r) = \text{softmax}_i(z(s, r)).
\]

\subsection{Proof of Lemma~\ref{lemma:2}}
\label{app:lemma2}
%更新为什么会影响到其他的边，连通图是怎样联通的
In Lemma~\ref{lemma:2}, we characterize the SFT process as adding edges between the connected subgraph of \( v_s \) and the connected subgraph of \( v_a \) in the LLM's knowledge graph. We now provide a proof of this statement.

\begin{proof}
We assume a standard cross-entropy loss on the output, and we perform a gradient update (SGD step) on the model parameters using the example $(s,r,a)$. Let $p_\theta(x \mid s,r)$ denote the model’s predicted probability for token $x$ as the answer given $(s,r)$. The cross-entropy loss for the correct answer $a$ is
\[
\mathcal{L} = -\log p_\theta(a \mid s, r).
\]
This loss pushes the model to increase the probability of $a$ while decreasing the probability of other tokens for the input $(s,r)$. 
Then, we examine the gradients with respect to various components. The gradient of $\mathcal{L}$ with respect to the hidden states for any token $x$ is
\begin{align*}
    \delta_h&=\frac{\partial \mathcal{L}}{\partial h(s,r)} \\
    &=\sum_{x} \frac{\partial \mathcal{L}}{\partial z_x(s,r)} \cdot \frac{\partial z_x(s,r)}{\partial h(s,r)}\\
    &= \sum_{x} 
(p_\theta(x \mid s, r) - \mathbb{I}\{x = a\}) \cdot U_{:,x}
\end{align*}
where $\mathbb{I}\{x=a\}$ is 1 for $x=a$ and 0 otherwise.$\delta_h$ points in the direction that increases the logit for $a$ and decreases logits for others.
Then, SGD update (with learning rate $\eta$) for $W^V$, and the new value vector for any other token $i \in V_s$ after updated is:
\begin{align*}
v_i^{\text{new}}&=(W^V+\Delta W^V)E[i]\\
&=W^V E[i]+\eta  \frac{\partial \mathcal{L}}{\partial W^V}\\ 
&= W^V E[i]+\eta \alpha_{s} \, \delta_h \, (E[s]^\top E[i])\\ 
& \quad + \eta \alpha_{r} \, \delta_h \, (E[r]^\top E[i])
\end{align*}
Because $E[i] \approx E[s]$, the inner product $E[s]^\top E[i]$ will be close to $|E[s]|^2$ (and $E[r]^\top E[i]$ is presumably small unless $r$ happened to be similar to $s$ in embedding). Thus $v_i$ gets a nearly identical adjustment in the $\delta_h$ direction.
For any token $i \in V_s$, consider its key after the update:
\begin{align*}
k_i^{\text{new}} 
&= (W^K+\Delta W^K)E[i]\\
&= W^K E[i] \\
&\quad + \eta \, \alpha_s (1 - \alpha_s) \cdot \delta_h^\top W^V (E[s] - E[r]) \\
&\quad \cdot q_r (E[s]^\top - E[r]^\top) E[i]
\end{align*}
$W^K$ is being adjusted so that the keys of $s$ and all similar tokens $i$ move closer in the direction of the relation’s query $q_r$. This increases $q_r \cdot k_i$ for each such $i$, thus increasing the attention weight $\alpha_{r\to i}$ when the model processes $(i,r)$ in the future.

The $W^Q$ update also specifically adjusted $q_r = W^Q E[r]$ to better align with $k_s$. This change benefits any input where the key is similar to $k_s$. In particular, $q_r$ will now have higher dot-product with $k_i$ for any $i$ in $V_s$ (since $k_i^{new} \approx k_s^{new}$). Thus, both $W^K$ and $W^Q$ updates reinforce the attention to any subject token similar to $s$. 

Now consider the forward pass for a new input $(i,r)$ after the update. The new hidden state for $(i,r)$ is then:
\begin{align*}
\alpha^{\text{new}}_i
&= \frac{
\exp\left( (q_r^{\text{new}})^\top k_i^{\text{new}} \right)
}{
\exp\left( (q_r^{\text{new}})^\top k_i^{\text{new}} \right) 
+ \exp\left( (q_r^{\text{new}})^\top k_r^{\text{new}} \right)
}
\end{align*}
Given our analysis, $(q_r^{new})^\top k_i^{new}$ is significantly larger than the old $(q_r^{old})^\top k_i^{old}$, and also larger relative to $(q_r^{new})^\top k_r^{new}$. Since the update was based on $s$ vs. $r$, we expect $(q_r^{new})^\top k_i^{new} \approx (q_r^{new})^\top k_s^{new}$ which was boosted. Thus $\alpha_{i}^{new}$ will be close to the $\alpha_{s}^{new}$ achieved for the training pair, which is likely near 1 if the model learned to almost fully attend to the subject. So the relation $r$ will heavily attend to $i$:
\begin{align*}
h(i,r)&\approx \alpha_i^{\text{new}}v_i^{\text{new}}+\alpha_r^{\text{new}}v_r^{\text{new}}  \\
&\approx v_i^{\text{new}}+(\text{small residual}).
\end{align*}
Because $v_i^{new}$ was updated to be nearly $v_s^{new}$ in the $\delta_h$ direction, and $v_s^{new}$ was tuned to align with $u_a$, it follows that $h(i,r)$ points toward $U_{:,a}$ as well. In other words, the hidden representation the model computes from $(i,r)$ is now oriented in a way that favors the answer $a$ and similar tokens.

Since $h(i,r) \approx v_i^{new}$ and $v_i^{new} \approx v_s^{new}$ and $v_s^{new}$ was pushed toward $U_{:,a}$, we have $z_a(i,r)$ greatly increased. The probabilities $P(x|i,r) = \mathrm{softmax}(z(i,r))$ will assign much more mass to $a$ and its neighbors. Therefore, the model’s predicted answer token $j = f_{[:,-1]}(i,r)$ will lie in the neighborhood $V_a$. Symbolically, $f_{[:,-1]}(i,r) = j$ with $j \in V_a$.
\end{proof}

\subsection{Proof of Theorem~\ref{theorem:gap}}
\label{app:proof_gap}
In this section, we prove Theorem~\ref{theorem:gap} and analyze why the unknown knowledge identified by few-shot prompting tends to correspond to nodes with lower degrees. Based on this observation, we further show that performing SFT on unknown knowledge results in a graph with fewer associated explicit edges, compared to the graph formed by fine-tuning on known knowledge.

\begin{assumption}
\label{ass:1}
We assume that, when using few-shot prompting, the attention mechanism guides the query \((s, r)\) to follow the patterns observed in the demonstrations \((s'_i, r, a'_i)\) when predicting the answer.
\end{assumption}

This assumption is reasonable based on prior work~\citep{brown2020languagemodelsfewshotlearners,pmlr-v202-von-oswald23a,akyrek2023what}, which demonstrates that language models can imitate demonstrated patterns via in-context learning.

\begin{proof}
In the transformer’s attention mechanism, the weight placed on any key–value pair is 

\[
\alpha_{t} = 
\frac{
\exp\left( (W^Q E[r])^\top (W^K E[t]) \right)
}{
\sum_{u \in \{s, r, s'_i, \dots\}} \exp\left( (W^Q E[r])^\top (W^K E[u]) \right)
}.
\]
Here, \( q = W^Q E[r] \) is the query vector for the relation token, and each key vector \( k_t = W^K E[t] \) corresponds to token \( t \). 

If a demonstration subject \( s'_i \) has an embedding \( E[s'_i] \) so close to \( E[s] \) that
\[
\| E[s'_i] - E[s] \| < \epsilon,
\]
then applying the same linear map \( W^K \) yields
\[
k_{s'_i} = W^K E[s'_i] \approx W^K E[s] = k_s.
\]
Because the two key vectors are nearly identical, their dot products with the query vector are also nearly the same:
\[
q^\top k_{s'_i} \approx q^\top k_s.
\]

According to Assumption~\ref{ass:1}, the prediction for \((s, r)\) follows the pattern established by the demonstrations \((s'_i, r, a'_i)\). Based on the derivation in Appendix~\ref{app:lemma2}, when
\[
q^\top k_{s'_i} \approx q^\top k_s,
\]
the resulting distribution \(p_\theta(x \mid \dots, s, r)\) will place most of its mass near \(a'_i\). In this case, if \(a \sim_\epsilon a'_i\), then the probability of correctly predicting the target answer \(a\) increases significantly.

Therefore, for known knowledge where few-shot prompting successfully leads to correct predictions, the pairs \((s, a)\) are typically close in the embedding space to many demonstration pairs \((s'_i, a'_i)\). This implies that \(|\mathcal{V}_s|\), the size of the similarity neighborhood in the constructed graph \(\mathcal{G}_s = (\mathcal{V}_s, \mathcal{E}^{\text{sim}})\), is relatively large. Similarly, \(|\mathcal{V}_a|\) is also larger.

According to Lemma~\ref{lemma:2}, this means that fine-tuning on known knowledge typically results in more than one new edge, while fine-tuning on unknown knowledge generally adds only one new edge. Thus, we have:
\[
|\mathcal{E}_{\text{kn}}| > |\mathcal{E}_{\text{unk}}|.
\]

Let \( D_{\text{test}} \) be a random sample of test triples \((s, r, a)\). Under the \emph{uniform-edge assumption} (i.e., every possible pair in \( \mathcal{V} \times \mathcal{V} \) is equally likely to be included in \( \mathcal{E}_r \)), the probability that a test triple \((s, r, a)\) is ``in'' the graph (i.e., can be answered correctly in one hop) is
\[
\Pr\left((v_s, v_a) \in \mathcal{E}_r\right) = \frac{|\mathcal{E}_r|}{|\mathcal{V}|^2}.
\]
Hence, the expected number of correctly answered test triples is
\[
|D_{\text{test}}| \times \frac{|\mathcal{E}_r|}{|\mathcal{V}|^2}.
\]

Define the \emph{factuality gap} between known- and unknown-fine-tuning as
\begin{align*}
\Delta_{\text{fact}} &= 
\left| \left\{ (v_s, v_a) \in E_{\text{kn}} \mid (s, r, a) \in D_{\text{test}} \right\} \right|\\
& \quad - \left| \left\{ (v_s, v_a) \in E_{\text{unk}} \mid (s, r, a) \in D_{\text{test}} \right\} \right|.
\end{align*}

Taking expectations under random sampling, we have:
\begin{align*}
\mathbb{E}[\Delta_{\text{fact}}] &= |D_{\text{test}}| \cdot \left( \frac{|\mathcal{E}_{\text{kn}}| - |\mathcal{E}_{\text{unk}}|}{|\mathcal{V}|^2} \right)\\
&\propto |\mathcal{E}_{\text{kn}}| - |\mathcal{E}_{\text{unk}}|\\
&> 0.
\end{align*}

That is,
\[
\Delta_{\text{fact}} \propto |\mathcal{E}_{\text{kn}}| - |\mathcal{E}_{\text{unk}}| > 0,
\]
which is exactly the statement of Theorem~\ref{theorem:gap}.

\end{proof}

\subsection{Proof of Theorem~\ref{theorem:ood}}
\label{app:proof_ood}
%语义距离越大，两个distribution子图的割集越小，ood受到的影响越小
We make several foundational proofs and attempt to provide a graph-theoretic analysis showing that the greater the semantic distance between the test set and the training set, the smaller the observed factuality gap on the test set.

We begin by proving the relationship between cosine similarity and the edge connectivity of the knowledge graph associated with the dataset.

\begin{proof}
First, we assume that all token embeddings are unit‐normalized, so for any two tokens \(i,j\)
\[||e_i||=||e_j||=1.\] 
Their Euclidean distance and cosine similarity are related, and under the neighborhood condition there is 
\[||e_i-e_j||<\epsilon \Longleftrightarrow \cos(e_i, e_j) > 1 - \frac{\epsilon^2}{2}.\] 

Let 
\[
\gamma = \mathbb{E}_{(s_{\text{test}}, s_{\text{train}})}\left[\cos\left(e_{s_{\text{test}}}, e_{s_{\text{train}}}\right)\right]
\]
be the average cosine similarity between a random test subject embedding and a random training subject embedding. Define the threshold
\[
\tau = 1 - \frac{\epsilon^2}{2}.
\]
Then, by Markov's inequality, the fraction of test–training pairs whose cosine exceeds \(\tau\) is bounded above by
\[
\Pr\left(\cos(e_{s_{\text{test}}}, e_{s_{\text{train}}}) > \tau \right) \leq \frac{\gamma}{\tau}.
\]
In particular, as \(\gamma\) decreases, so does the probability that a random test subject lies within an \(\epsilon\)-ball of a random training subject. The same argument applies to object embeddings \(a\).

Since embedding neighborhoods are \textbf{independent} for the subject and the object, the joint probability that a given training triple implants the correct test edge is bounded by
\begin{align*}
&\Pr\left( s_{\text{test}} \in \mathcal{V}_{s_{\text{train}}} \;\wedge\; a_{\text{test}} \in \mathcal{V}_{a_{\text{train}}} \right)\\
&= \Pr\left( \cos(e_{s_{\text{test}}}, e_{s_{\text{train}}}) > \tau \right)\\
& \quad \quad \times \Pr\left( \cos(e_{a_{\text{test}}}, e_{a_{\text{train}}}) > \tau \right)\\
&\leq \left( \frac{\gamma}{\tau} \right)^2.
\end{align*}

Thus, each training triple contributes to the test‐set edge‐coverage only with probability at most \(\left( \frac{\gamma}{\tau} \right)^2\).
Then the factuality gap scales at most like
\[
\left( \frac{\gamma}{\tau} \right)^2 \cdot D_{\text{kn}} 
- \left( \frac{\gamma}{\tau} \right)^2 \cdot D_{\text{unk}} 
= \left( \frac{\gamma}{\tau} \right)^2 \cdot \left( D_{\text{kn}} - D_{\text{unk}} \right).
\]

In particular, as the average test–train cosine similarity \(\gamma\) decreases, the factor 
\(\left( \frac{\gamma}{\tau} \right)^2\) becomes smaller, thereby reducing the factuality gap proportionally.

\end{proof}

\subsection{Proof of Theorem~\ref{theorem:icl}}
\label{app:proof_icl}
%icl作用，利用transformer性质，prompt graph连通图影响模型主图
In this work, the term ICL prompt refers to two specific types of prompts: few-shot prompts and CoT prompts. In the following, we consider these two types separately to provide the theoretical analysis.

\begin{proof}
    
Let \( \Pi =\{(s'_i,r,a'_i)\}\) be the few-shot prompt that provided to the LLM together with the input pair \( (s, r) \). This prompt  can be interpreted as an auxiliary knowledge graph \( \mathcal{G}_{\Pi}=(\mathcal{V}_{\Pi},\mathcal{E}_{r_{\Pi}},\mathcal{E}^{\text{sim}}_{\Pi})\). The graph includes not only the triples \((s, r, a)\) from the demonstrations, but also the edges connecting them to semantically similar nodes that are implicitly related within the same domain.

With graph \( \mathcal{G}_{\Pi}\), the updated knowledge graph becomes
\[
\mathcal{G}^\star = \mathcal{G}_{\text{unk}/\text{kn}} \cup \mathcal{G}_{\Pi}.
\]

Since \( \mathcal{G}_{\Pi}, \mathcal{G}_{\text{unk}}, \mathcal{G}_{\text{kn}} \subseteq \mathcal{G}_r \), with a knowledge prompt that has enough semantic connection with the in-distribution data, there exists a sufficiently large subgraph \( \mathcal{G}_{\mathcal{P}} \) such that
\[
|\mathcal{E}_{\Pi} \cap \mathcal{E}_{\text{kn}}| > |\mathcal{E}_{\Pi} \cap \mathcal{E}_{\text{unk}}|.
\]

Then, the factuality gap is
\begin{align*}
\Delta_{\text{fact}}^\star 
&= \lambda\left(|\mathcal{E}_{\text{kn}} \cup \mathcal{E}_{\mathcal{P}}| - |\mathcal{E}_{\text{unk}} \cup \mathcal{E}_{\mathcal{P}}|\right) \\
&= \lambda\left[\left(|\mathcal{E}_{\text{kn}}| + |\mathcal{E}_{\mathcal{P}}| - |\mathcal{E}_{\text{kn}} \cap \mathcal{E}_{\mathcal{P}}|\right) \right. \\
&\quad \left. - \left(|\mathcal{E}_{\text{unk}}| + |\mathcal{E}_{\mathcal{P}}| - |\mathcal{E}_{\text{unk}} \cap \mathcal{E}_{\mathcal{P}}|\right)\right] \\
&= \lambda(|\mathcal{E}_{\text{kn}}| - |\mathcal{E}_{\text{unk}}|\\
&\quad \quad -\left(|\mathcal{E}_{\text{kn}} \cap \mathcal{E}_{\mathcal{P}}| - |\mathcal{E}_{\text{unk}} \cap \mathcal{E}_{\mathcal{P}}|\right)) \\
&< \lambda\left(|\mathcal{E}_{\text{kn}}| - |\mathcal{E}_{\text{unk}}|\right) \\
&= \Delta_{\text{fact}}.
\end{align*}
According to Appendix~\ref{app:proof_gap}, there \(\lambda = \frac{|\mathcal{D}_{\text{test}}|}{|\mathcal{V}|^2}\). Therefore, we can get
\[\Delta_{\text{fact}}^\star < \Delta_{\text{fact}}.\]

Let \( C =(s,r_1,a_1,r_2,a_2,\dots,r_k,a)\) be the CoT prompt that provided to the LLM together with the input pair \( (s, r) \). This prompt  can be interpreted as an auxiliary knowledge graph \( \mathcal{G}_{C}=(\mathcal{V}_{C},\mathcal{E}_{C})\). The graph consists of the complete set of nodes and edges that lie along the reasoning path from the subject \(s\) to the object \(a\).

With graph \( \mathcal{G}_{\Pi}\), the updated knowledge graph becomes
\[
\mathcal{G}_{\text{unk}/\text{kn}}^\star = \mathcal{G}_{\text{unk}/\text{kn}} \cup \mathcal{G}_{C}.
\]

The new factuality gap is defined as
\[
\Delta_{\text{fact}}^\star = 
\left| \{ \text{covered by } \mathcal{G}_{\text{kn}}^\star \} \right| - 
\left| \{ \text{covered by } \mathcal{G}_{\text{unk}}^\star \} \right|.
\]

But for every test triple \((s, r, a)\) that is explained by the CoT prompt, it is covered by both augmented graphs. Therefore, its contribution to the gap is \(1 - 1 = 0\). Any remaining gap can only come from test triples not supported by CoT.

In the extreme case where CoT covers the entire test set, we have:
\[
\Delta_{\text{fact}}^\star = 0.
\]

More generally, since the same CoT subgraph is added to both graphs, the only remaining difference in coverage comes from test triples outside the scope of CoT. Thus, we have:
\[
\Delta_{\text{fact}}^\star \leq \Delta_{\text{fact}}.
\]

\end{proof}

\section{Experiment Details}
\label{app:exp}
\subsection{QA tasks} \label{app:qa}
\paragraph{Data processing.} 

For the Entity Questions task, we adopt the experimental framework outlined by \citet{gekhman-etal-2024-fine}. Specifically, we select train split and dev split data from the following relation subsets: \texttt{P131}, \texttt{P136}, \texttt{P17}, \texttt{P19}, \texttt{P26}, \texttt{P264}, \texttt{P36}, \texttt{P40}, \texttt{P495}, \texttt{P69}, \texttt{P740}, and \texttt{P800} for both training and evaluation purposes. The remaining relation subsets are reserved for out-of-distribution (OOD) testing, as described in Section \ref{sec:ood}. 
We employ a few-shot learning approach to classify the Unknown and Known datasets. Within the dev split, we randomly select 10 sets, each containing 4 examples, and apply both greedy and random sampling decoding methods. For random sampling, the following parameters are used: \texttt{temperature}=0.5, \texttt{top\_p}=1.0, \texttt{top\_k}=40, and 16 answers are sampled. The data is classified as either Unknown or Known based on the accuracy of the greedy search and random sample. If at least one correct answer is obtained from either the greedy search or random sampling, the data is classified as Known. We perform this filtering procedure for each relation subset and subsequently use the filtered Unknown and Known splits to balance the data across categories. For each relation, we take the smaller data size between the Known split and the Unknown split as the final data size, in order to ensure that the Known and Unknown splits have equal amounts of data under each relation.
 After filtering, the number of Unknown and Known samples for each of the four models is as follows: Llama Base: 28,337, Llama Instruct: 31,226, Mistral Base: 30,952, and Mistral Instruct: 31,335. For evaluation, we randomly select 2,000 samples from the development dataset corresponding to the relation subsets used in the training dataset.

For PopQA, similar to Entity Questions, we perform the splitting for each question type individually. First, each subclass dataset is randomly divided into a training set and an evaluation set in a 4:1 ratio. Then, the training set is further split into two halves to ensure an equal distribution of each type of question. We also use few-shot prompting to filter the Unknown and Known splits. The difference is that, considering the smaller size of the PopQA dataset, we randomly select only 3 few-shot groups from the evaluation set, while keeping the other filtering parameters consistent with those used for Entity Questions. Finally, the number of Known and Unknown samples used for each of the four models is as follows: LLaMA Base: 3,659; LLaMA Instruct: 3,589; Mistral Base: 3,488; and Mistral Instruct: 3,421. The evaluation dataset consists of 2,858 samples.

% 对于MMLU，我们仍然采用few-shot learning的方法，不过进行了一些简化，我们直接选取MMLU的dev split的5个数据作为一组fewshot examples, 除了random sample的数量改为4之外，其他模型超参数的设置和Entity Questions相同。我们选取MMLU中的test split作为训练数据，而val split作为evaluation的数据。对于训练数据，同样确保每一类Ukonwn和Known的长度取数据量最小的那个。最后，四种模型的Unkonwn和Known的数据量分别为：Llama Base: 2724 Llama Instruct: 2730 Mistal Base: 2994 Mistral Instruct: 4128. evaluation dataset的长度为1531.
For MMLU, we also adopt a few-shot learning approach, but with some simplifications. We directly select 5 data points from the MMLU dev split as a group of few-shot examples. Apart from changing the number of random samples to 4, the other model hyperparameters are set the same as in Entity Questions. We use the test split of MMLU as the training data and the val split as the evaluation data. For the training data, we ensure that the Unknown and Known datasets have the same number of samples by taking the smaller size from each class. Finally, the number of Unknown and Known samples for the four models is as follows: Llama Base: 2,724, Llama Instruct: 2,730, Mistral Base: 2,994, Mistral Instruct: 4,128. The length of the evaluation dataset is 1,531.

\paragraph{Training Details.}
% 我们根据数据集和模型将所有训练分成12组，每组又包含unkonwn,konwn,mixed三种数据子集上的训练，我们确保在组内子集训练参数完全一致。

% 三个数据集的训练超参数中，batch size=128，我们采用fixed learning rate，其中Llama Base和Llama Instruct两个模型的learning rate为1e-5，Mistral Base和Mistal Instruct两个模型在Entity Questions上的learning rate为5e-6,其余数据集为1e-6。训练没有使用其他的正则化训练方法。 三个数据集的训练都是以evaluation set上的accuracy最好的模型作为early stop的模型，训练完全部epoch确保其loss收敛的模型为Convergence模型。在Entity Questions dataset上，所有模型都训练了20个epoch. 在PopQA上Llama系列模型训练了15个epcoh,Mistral Base和Mistral Instruct上分别训练30和35个epoch.在MMLU上,Llama系列模型训练了15个epoch,在Mistal系列模型上训练了30个epoch.

% 此外，对于SFT过程的prompt,PopQA数据集上完全采用原始问题和答案，Entity Questions的问题prompt形式如下：

We divide all the training into 12 groups based on the 3 datasets and 4 models, with each group containing training on the Unknown and Known subsets. We ensure that the training parameters are exactly the same within each group.

For all the 12 groups, the training hyperparameters are set as follows: the batch size is 128, and we use a fixed learning rate. Specifically, the learning rates for Llama Base and Llama Instruct are set to 1e-5, while for Mistral Base and Mistral Instruct, the learning rate for Entity Questions is 5e-6, and for the other datasets, it is set to 1e-6. No additional regularization methods are used during training. The training for all 12 groups uses the model with the best accuracy on the evaluation set as the Early Stop model, and the model whose loss converged after completing all epochs is considered the Convergence model. 

For the Entity Questions and PopQA dataset, all models are trained for 20 epochs. For MMLU, the Llama models are trained for 15 epochs, while the Mistral models are trained for 30 epochs. All of the models are trained on an 8× RTX 6000 Ada Generation 48G setup.

Additionally, for the SFT process prompt, the PopQA dataset use the original questions and answers, while the question prompt format for the Entity Questions dataset is as follows:
\begin{tcolorbox}[
    colframe=black, % Frame color
    colback=gray!15,  % Background color
    coltitle=black, % Title color
    fonttitle=\bfseries, % Bold title font
    rounded corners, % Sharp corners for the box
    enhanced, % Additional styling options
    left=6pt, right=6pt, top=6pt, bottom=6pt, % Padding
    boxrule=1pt, % Border thickness
    arc=6pt,
    width=\linewidth % Box width
]
Answer the following question.\textbackslash n Who is Caitlin Thomas married to?
\end{tcolorbox}
% MMLU的问题的prompt形式如下：

The question prompt format for the MMLU dataset is as follows:

\begin{tcolorbox}[
    colframe=black, % Frame color
    colback=gray!15,  % Background color
    coltitle=black, % Title color
    fonttitle=\bfseries, % Bold title font
    rounded corners, % Sharp corners for the box
    enhanced, % Additional styling options
    left=6pt, right=6pt, top=6pt, bottom=6pt, % Padding
    boxrule=1pt, % Border thickness
    arc=6pt,
    width=\linewidth % Box width
]
The following is a multiple choice question, paired with choices. Answer the question in format: 'Choice:content'.\textbackslash n\textbackslash n\#\#\# Question:\textbackslash nThe cyclic subgroup of Z\_24 generated by 18 has order\textbackslash n\textbackslash n\#\#\#  Choices:\textbackslash nA) 0 B) 4 C) 2 D) 6 \textbackslash n\textbackslash n\#\#\#  Answer:\textbackslash n
\end{tcolorbox}

\paragraph{Evaluation Details.}
% 我们Exact Match作为metric来衡量模型的evaluation accuracy, 测试时的问题的prompt形式和训练时相同，测试时的模型采用greedy search的decoding方式，max\_token=10
We use Exact Match as the metric to measure the model's evaluation accuracy. During testing, the prompt format of the questions is the same as during training. The model during testing uses the greedy search decoding method with a \texttt{max\_new\_token} value of 10.

\subsection{Open-ended generation tasks} \label{app:gen}
\paragraph{Data processing.}
We utilize the WikiBios~\citep{kang2024unfamiliar} data directly, randomly selecting 2,000 entries as the training set and 500 entries as the evaluation dataset. For the training set partition, we also employ a few-shot learning approach. In the evaluation set, we select 4 examples and used the random sample decoding method to sample two answers, with \texttt{max\_token=32}. The remaining decoding parameters are the same as in Entity Questions. To assess the accuracy of the answers, we employed the \textbf{FActScore} metric. The GPT model used for this task  is \texttt{gpt-3.5-turbo-0125}, with raw scores and no penalties applied for the \texttt{num\_fact} parameter. Each data point is evaluated individually, and the average of the two sampled answers is taken. Based on the resulting FActScore, the training set is then divided into two parts: the higher-scoring subset is classified as Known, while the lower-scoring subset is classified as Unknown.

% 我们直接使用\citet{kang2024unfamiliar}中的WikiBios的数据，从中随机选取了2000条作为训练集，500条作为evaluation dataset, 对于训练集的划分，我们也是采用few-shot learning的方法，在eval set中选4个examples，采用random sample的decoding的方式，从中sample出了两个答案，max\_token=32,其余decoding参数和上面Entity Questions一致。我们使用FActScore作为metric来判断答案的正确性。这里采用的GPT model为gpt-3.5-turbo-0125，用原始的score不考虑针对num\_fact的惩罚。我们逐条数据测试，取两个答案的平均值，最后根据FActScore将训练集分成两部分，高的那部分为Known,低的那部分为Unknown.
\paragraph{Training Details.}
% 该数据集只在Llama Base和Mistral Base上训练，其中batch size=128, fixed learning rate为1e-5, 没有采用其他正则化方法。训练到loss收敛到0.01以内停止，作为Convergence Model,选取 eval loss最低的model为 early stop对应的model 

The dataset is trained only on Llama Base and Mistral Base, with a batch size of 128 and a fixed learning rate of 1e-5. No additional regularization methods are used. Training stops when the loss converged to below 0.01, and this model is considered the Convergence Model. The model with the lowest evaluation loss is selected as the early stop model.

\paragraph{Evaluation Details.}
We used \textbf{FActScore} as the evaluation metric, with the same data processing settings as described above.

\subsection{Toy Example} \label{app:toy}
For our Toy Example, we utilized the Llama3.3-70B-Instruct\footnote{\url{https://huggingface.co/meta-llama/Llama-3.3-70B-Instruct}} model, incorporating data sampled from the EntityQuestions dataset.

\textbf{Data processing.} We employ the Llama3.3-70B model to construct the \textit{Known} knowledge set by querying the model with the original questions. To each question, we append the phrase \textit{"Answer the following question."} before the question itself to form a complete query, without relying on additional few-shot examples. Specifically, we apply a greedy sampling method, limiting the model's output to a maximum of 10 tokens, and verified whether the ground truth answer is present in the model's response. If the ground truth answer is included, we identifiy the subject words in the question. For each subject word longer than two letters, we introduce a fixed perturbation, "\$\&". For subject words of three letters, the perturbation is inserted after the first letter. For subject words longer than three letters, the perturbation is applied before the second letter. The modified question is then re-entered into the model to ensure that the resulting response did not contain the answer to the original question, and regarded as the \textit{Unknown knowledge}.

Below is an example of our known and unknown set consturction, using the real question from relation \texttt{P26}. The question in this case is “Who is Caitlin Thomas married to?”, and the ground truth answer is “Dylan Thomas”. The subject words in the question is “Caitlin Thomas”.

\begin{tcolorbox}[
    colframe=black, % Frame color
    colback=gray!15,  % Background color
    coltitle=black, % Title color
    fonttitle=\bfseries, % Bold title font
    rounded corners, % Sharp corners for the box
    enhanced, % Additional styling options
    left=6pt, right=6pt, top=6pt, bottom=6pt, % Padding
    boxrule=1pt, % Border thickness
    arc=6pt,
    width=\linewidth % Box width
]
Q: Answer the following question.\textbackslash n Who is Caitlin Thomas married to?\\
A: Caitlin Thomas. \\
Modified: Answer the following question.\textbackslash n Who is C\$\&aitl\$\&in T\$\&hom\$\&as married to?\\
A: Rio de Janeiro. 
\label{box:cot}
\end{tcolorbox}

We combine the following relations from the EntityQuestion dataset: \texttt{P131}, \texttt{P136}, \texttt{P17}, \texttt{P19}, \texttt{P26}, \texttt{P264}, \texttt{P36}, \texttt{P40}, \texttt{P495}, \texttt{P69}, \texttt{P740}, and \texttt{P800}, resulting in a training set of 2,000 data entries and a test set of 1,000 for the \textit{Known} and \textit{Unknown}  dataset.

\textbf{Training Details.} During the training of the Toy Example, we use a learning rate of 2e-5, a batch size of 128, and a weight decay of 0. We apply a cosine learning rate scheduler with a warm-up of 64 steps. We use the training data template detailed in Appendix~\ref{app:qa}, and trained the model for a total of 50 epochs on an 8$\times$6000 Ada 48G setup.

\textbf{Toy Example CoT prompt.} To mitigate the performance gap caused by fine-tuning on different data filters, we employ the following Chain-of-Thought (CoT) prompt to guide the model in reasoning and answering the questions.
\begin{tcolorbox}[
    colframe=black, % Frame color
    colback=gray!15,  % Background color
    coltitle=black, % Title color
    fonttitle=\bfseries, % Bold title font
    rounded corners, % Sharp corners for the box
    enhanced, % Additional styling options
    left=6pt, right=6pt, top=6pt, bottom=6pt, % Padding
    boxrule=1pt, % Border thickness
    arc=6pt,
    width=\linewidth % Box width
]
Ignore all the special characters in the following question. Think step by step. First, clean all special characters in the question. In this step, you might see some unicode characters in foreign languages. Next, rethink the cleaned question. Finally, give the detailed answer of the cleaned question with short explanation. 
\label{box:cot}
\end{tcolorbox}

\subsection{OOD Generalization} \label{app:ood}
For near in-distribution tasks, We follow \citet{gekhman-etal-2024-fine} and sample non-overlapping data from the remaining relation subsets of the Entity Questions  with 3000 data points to create near in-distribution test set \texttt{eq\_ood}.We use the entire PopQA evaluation dataset as  near in-distribution test sets \texttt{pop\_ood}. The cosine similarities between \texttt{eq\_ood}, \texttt{pop\_ood}, and the ID test set are 0.86 and 0.82, respectively. For the open-world task, we choose MMLU, which provides more diverse data and significantly different question formats. We select 50 samples from each of the 57 MMLU tasks to create a complete \texttt{mmlu\_ood} set. After embedding, the cosine similarity between \texttt{mmlu\_ood} and the ID test set is 0.55.

\subsection{Finetuned on Small Dataset}
\label{app:small}
In this section, we introduce the experimental setup for evaluating the factuality gap resulting from fine-tuning on a small subset versus the whole dataset. The construction of the PopQA training and evaluation sets has already been described in Appendix~\ref{app:exp}. We fine-tune both the LLaMA Base and Mistral Base models on two dataset settings: (1) the full PopQA training set, consisting of 11,409 samples and (2) a randomly selected 5\% subset of the full data, consisting of 561 samples. The training hyperparameters follow those specified in Appendix~\ref{app:exp} for the corresponding PopQA experiments. The only difference is that here we train for 10 epochs and select the final model based on early stopping.

The evaluation settings remain the same as in previous experiments, including the reuse of the original ICL prompt design.

% \section{Train Acc Curves} \label{app:train_curve}

% % QA tasks 所有的训练过程的training acc曲线如图\ref{fig:train_acc}所示，generation task训练过程的training loss曲线如图\ref{fig:train_loss}所示

% The training accuracy curve for all QA tasks is shown in Figure \ref{fig:train_acc}, while the training loss curve for the generation task is shown in Figure \ref{fig:train_loss}.

% \begin{figure}[h]
%   \includegraphics[width=\linewidth]{figures/train_loss.pdf}
%   \caption{Training loss of generation task}
%   \label{fig:train_loss}
% \end{figure}

% \begin{figure*}[t]
%   \includegraphics[width=\linewidth]{figures/train_acc.pdf}
%   \caption{Training accuracy of QA tasks}
%   \label{fig:train_acc}
% \end{figure*}

\section{Prompt Design Details} \label{app:cot}
% few-shot选取的是Known split中的examples, 考虑到example的长度和有效性，其中PopQA和Entity Questions中选4个,MMLU中选3个。我们使用GPT-4o来生成各类任务的CoT prompt, 每个数据集都分问题类型输入few-shot learning的examples和生成CoT的instruction,从而得到对应问题类型的few-shot CoT prompt.各个数据集的instruction如下：
For few-shot learning, we select examples from the Known split. Considering the length and effectiveness of the examples, 4 examples were selected from PopQA and Entity Questions, while 3 examples were selected from MMLU. We used GPT-4 to generate the CoT prompts for each type of task. For each dataset, we input the few-shot learning examples and generate the CoT instructions according to the question type, thus obtaining the corresponding few-shot CoT prompt for each question type. The instructions for each dataset are as follows:

\begin{tcolorbox}[
    colframe=black, % Frame color
    colback=gray!15,  % Background color
    coltitle=black, % Title color
    fonttitle=\bfseries, % Bold title font
    rounded corners, % Sharp corners for the box
    enhanced, % Additional styling options
    left=6pt, right=6pt, top=6pt, bottom=6pt, % Padding
    boxrule=1pt, % Border thickness
    arc=6pt,
    width=\linewidth % Box width
]
\textbf{Entity Questions, PopQA:}
Follow the few shot Chain of Thought example format: Question:\{\} Analysis:\{\} Answer:\{\} to modify the format and generate analysis \
                 of the entity in each question of the QA pairs below.\
                 The analysis should describe the related information of the entity shortly in the question in order to lead to the answer:
\end{tcolorbox}

\begin{tcolorbox}[
    colframe=black, % Frame color
    colback=gray!15,  % Background color
    coltitle=black, % Title color
    fonttitle=\bfseries, % Bold title font
    rounded corners, % Sharp corners for the box
    enhanced, % Additional styling options
    left=6pt, right=6pt, top=6pt, bottom=6pt, % Padding
    boxrule=1pt, % Border thickness
    arc=6pt,
    width=\linewidth % Box width
]
\textbf{MMLU:}
'Follow the few-shot Chain of Thought example format: Question:\{\} Choices:\{\} Analysis:\{\} Answer:\{\} to modify the format and generate analysis of the critical entity in each multiple choice question below. The analysis should describe the related information of the entity in the question shortly in order to lead to the answer:\textbackslash n
\end{tcolorbox}

\section{Abalation Study Details} \label{app:aba}

% \begin{table}[h]
% \setlength{\tabcolsep}{1pt}
% \begin{tabular}{ccccc}
% \hline
%          & Llama & Llama-Ins & Mistral & Mistral-Ins \\ \hline
% EQ       & 35.65 & 32.90          & 30.25   & 29.70            \\ 
% PopQA    & 36.39 & 32.16          & 32.86   & 29.60            \\ 
% MMLU     & 59.70 & 61.85          & 50.42   & 55.72            \\ 
% WikiBios & 57.53 &                & 62.14   &                  \\ \hline
% \end{tabular}
% \caption{Few-shot learning on pre-trained models}
% \label{tab:pretrain}
% \end{table}
 
% 对于few-shot learning的examples的选择，\ref{tab}展示了所有Unkonwn examples的测试结果，对于Table\ref{tab:fewshot_only}中只用Known examples的结果可以发现用Known examples的大部分模型的factuality都有提升，
For the selection of few-shot learning examples, Table~\ref{tab:unknown_fewshot} shows the test results for all \textit{Unknown} examples. The testing of \textit{Unknown} examples is the same as for \textit{Known} examples, where 3 sets are randomly selected from the corresponding dataset, with each set containing 4 examples. The set with the best performance is then chosen. As for the results using only Known examples in Table \ref{tab:known_fewshot}, it can be observed that for most models, the factuality improves when using Known examples.

% 对于CoT的消融实验，只用few-shot和加上CoT的实验结果如Table\ref{tab:fewshot_only}和Table\ref{tab:cot_only},从中可以对比加上CoT和不加之间的区别。我们可以发现在PopQA和Entity Questions上训练的模型的factuality提升，而MMLU则结果不稳定，有时不会比没有CoT时提升。我们猜想这可能是因为CoT使得文本过长导致的性能下降
For the ablation experiment of CoT, the results using only few-shot learning and those with the addition of CoT are shown in Table \ref{tab:known_fewshot} and Table \ref{tab:cot_fewshot}, respectively. By comparing the results, we can observe the differences between the models with and without CoT. We find that the factuality of the models trained on PopQA and Entity Questions improves, while the results on MMLU are more unstable and sometimes do not show any improvement with the addition of CoT. We hypothesize that this may be due to CoT causing the text to become too long, leading to a performance degradation.

% 对于问题format变化的消融实验，我们使用GPT-4o来进行Entity Questions中的evaluation dataset中2000条数据进行3次问题重述，3次重述的instruction分别为
For the ablation experiment on the variation of question formats, we used GPT-4 to rephrase 2,000 data points from the Entity Questions evaluation dataset three times. The instructions for the three rephrasings are as follows:
\begin{tcolorbox}[
    colframe=black, % Frame color
    colback=gray!15,  % Background color
    coltitle=black, % Title color
    fonttitle=\bfseries, % Bold title font
    rounded corners, % Sharp corners for the box
    enhanced, % Additional styling options
    left=6pt, right=6pt, top=6pt, bottom=6pt, % Padding
    boxrule=1pt, % Border thickness
    arc=6pt,
    width=\linewidth % Box width
]
Please rephrase this question with Minor Difference. Just return the rephrased question without additional word.

Please rephrase this question with Moderate Difference. Just return the rephrased question without additional word.

 Please rephrase this question with Radical Difference. Just return the rephrased question without additional word.
\end{tcolorbox}

\begin{table*}[]
\begin{tabular}{cccccccccc}
\hline
\multicolumn{1}{c}{\multirow{2}{*}{Benchmark}} & \multirow{2}{*}{Split} & \multicolumn{2}{c}{Llama} & \multicolumn{2}{c}{Llama-Instruct} & \multicolumn{2}{c}{Mistral} & \multicolumn{2}{c}{Mistral-Instruct} \\ \cline{3-10} 
\multicolumn{1}{c}{}                           &                        & ES          & Con.        & ES               & Con.            & ES           & Con.         & ES                & Con.             \\ \hline
\multirow{2}{*}{EQ}                            & Unknow                 & 36.60       & 29.60       & 33.25            & 26.20           & 26.55        & 18.50        & 30.95             & 19.20            \\
                                               & Known                  & 41.45       & 39.55       & 39.45            & 37.55           & 33.80        & 33.75        & 32.55             & 32.80            \\ \hline
\multirow{2}{*}{PopQA}                         & Unknown                & 32.61       & 29.46       & 27.64            & 26.77          & 28.87        & 28.45        & 29.01             & 28.20            \\
                                               & Known                  & 35.97       & 34.71       & 32.68           & 31.18           & 31.39        & 31.42        & 30.06             & 29.92            \\ \hline
\multirow{2}{*}{MMLU}                          & Unknown                & 54.02       & 53.43       & 64.34            & 64.14           & 54.02        & 53.63        & 55.26             & 55.45            \\
                                               & Known                  & 66.62       & 66.69       & 66.95            & 66.75           & 56.89        & 57.09        & 59.70             & 59.96            \\ \hline 
\multirow{2}{*}{WikiBios}                      & Unknown                & 54.18       & 48.62       &            &           & 48.24        & 38.18        &                   &                  \\
                                               & Known                  & 54.81       & 50.63       &                  &                 & 48.54        & 36.48        &                   &                  \\ \hline
\end{tabular}
\caption{Few-shot learning with Unknown examples}
\label{tab:unknown_fewshot}
\end{table*}

% \usepackage{multirow}

% \usepackage{multirow}
\begin{table*}[]
\begin{tabular}{cccccccccc}
\hline
\multirow{2}{*}{Benchmark} & \multirow{2}{*}{Split} & \multicolumn{2}{c}{Llama} & \multicolumn{2}{c}{Llama-Instruct} & \multicolumn{2}{c}{Mistral} & \multicolumn{2}{c}{Mistral-Instruct} \\ \cline{3-10} 
                           &                        & ES          & Con.        & ES               & Con.            & ES           & Con.         & ES                & Con.             \\ \hline
\multirow{2}{*}{EQ}        & Unknow                 & 39.10       & 32.10       & 37.65            & 34.40           & 31.70        & 25.05        & 32.05             & 21.25            \\
                           & Known                  & 41.75       & 39.90       & 39.80            & 37.80           & 31.40        & 30.15        & 33.05             & 33.90            \\ \hline
\multirow{2}{*}{PopQA}     & Unknown                & 37.05       & 33.80       & 31.07           & 28.52           & 28.97        & 28.55        & 29.25            & 28.38            \\
                           & Known                  & 36.91       & 36.00       & 34.29            & 33.00           & 31.42        & 31.60        & 30.27             & 30.13            \\ \hline
\multirow{2}{*}{MMLU}      & Unknown                & 54.80       & 54.60       & 64.99            & 65.32           & 55.39        & 55.13        & 56.24             & 56.43            \\
                           & Known                  & 67.60       & 67.86       & 69.30            & 68.84           & 58.46        & 58.39        & 60.48             & 60.74            \\ \hline
\multirow{2}{*}{WikiBios}  & Unknown                & 53.72       & 47.03       &                  &                 & 47.93        & 35.53        &                   &                  \\
                           & Known                  & 55.61       & 50.09       &                  &                 & 50.58        & 38.97        &                   &                  \\ \hline
\end{tabular}
\caption{Few-shot learning with Known examples}
\label{tab:known_fewshot}
\end{table*}

\begin{table*}[h]
\begin{tabular}{cccccccccc}
\hline
\multicolumn{1}{c}{\multirow{2}{*}{Benchmark}} & \multirow{2}{*}{Split} & \multicolumn{2}{c}{Llama} & \multicolumn{2}{c}{Llama-Instruct} & \multicolumn{2}{c}{Mistral} & \multicolumn{2}{l}{Mistral-Instruct} \\ \cline{3-10} 
\multicolumn{1}{c}{}                           &                        & ES          & Con.        & ES               & Con.            & ES           & Con.         & ES                & Con.             \\ \hline
\multirow{2}{*}{EQ}                            & Unknow                 & 41.55       & 38.95       & 41.00            & 37.40           & 35.35        & 32.95        & 35.25             & 30.05            \\
                                               & Known                  & 43.45       & 42.20       & 41.20            & 40.70           & 38.25        & 37.95        & 33.15             & 32.65            \\ \hline
\multirow{2}{*}{PopQA}                         & Unknown                & 39.82       & 37.89      & 35.06            & 34.00           & 35.93        & 35.76        & 31.46             & 31.32            \\
                                               & Known                  & 38.77       & 38.66       & 35.55            & 36.18           & 35.93        & 35.90        & 31.63             & 31.84            \\ \hline
\multirow{2}{*}{MMLU}                          & Unknown                & 45.79       & 47.35       & 64.34            & 64.01           & 53.04        & 53.49        & 58.00             & 60.09            \\
                                               & Known                  & 56.56       & 56.83       & 65.12            & 65.45           & 56.50        & 58.13        & 61.07             & 60.94            \\ \hline
\end{tabular}
\caption{Few-shot learning with CoT}
\label{tab:cot_fewshot}
\end{table*}

% \section{Attention Visualization} \label{app:vis}
% % 这里补充了两个问题来进行三种情况下的attention可视化，两个问题分别是：‘Which country is Valea Coacăzei River located in?’ 和 ‘Where was Margaret Mwanakatwe born?’，atention map分别在\ref{fig:ft_model2}和\ref{fig:ft_model3}中
% Two additional questions are added to visualize attention in three different cases. The two questions are: "Which country is Valea Coacăzei River located in?" and "Where was Margaret Mwanakatwe born?". The attention maps are shown in Figures \ref{fig:ft_model2} and \ref{fig:ft_model3}, respectively.

% \begin{figure*}[h]
% \centering
%   \includegraphics[width=\linewidth]{figures/ft_model2.pdf}
%    \includegraphics[width=\linewidth]{figures/ft_model_fewshot2.pdf}
%     \includegraphics[width=\linewidth]{figures/ft_model_cot2.pdf}
%   \caption{Attention maps of fine-tuned models. Top: Origin prompt. Middle: With few-shot learning Bottom: With CoT. Left: Fine-tuned on \textit{Unknown} data. Right: Fine-tuned on \textit{Known} data. Subject entity is "Valea Coacăzei River".}
%   \label{fig:ft_model2}
% \end{figure*}

% \begin{figure*}[h]
% \centering
%   \includegraphics[width=\linewidth]{figures/ft_model3.pdf}
%    \includegraphics[width=\linewidth]{figures/ft_model_fewshot3.pdf}
%     \includegraphics[width=\linewidth]{figures/ft_model_cot3.pdf}
%   \caption{Attention maps of fine-tuned models. Top: Origin prompt. Middle: With few-shot learning Bottom: With CoT. Left: Fine-tuned on \textit{Unknown} data. Right: Fine-tuned on \textit{Known} data. Subject entity is "Margaret Mwanakatwe".}
%   \label{fig:ft_model3}
% \end{figure*}

\end{document}